\documentclass{article}

\PassOptionsToPackage{numbers, compress}{natbib}

\usepackage[final]{neurips_2019}

\usepackage[utf8]{inputenc} 
\usepackage[T1]{fontenc}    
\usepackage{hyperref}       
\usepackage{url}            
\usepackage{booktabs}       
\usepackage{amsfonts}       
\usepackage{nicefrac}       
\usepackage{microtype}      

\usepackage{lscape} 
\usepackage{caption}
\def\name{Bidirectional-Inference Variational Autoencoder}
\def\nm{BIVA\xspace}

\usepackage{subcaption}
\usepackage{amsmath,amsfonts,amsthm,amssymb} 
\usepackage{booktabs}
\usepackage{wrapfig}
\usepackage[normalem]{ulem}
\usepackage{amsmath}
\usepackage[]{todonotes}
\usepackage{tikz}
\usepackage{xspace}
\usetikzlibrary{bayesnet}
\usetikzlibrary{arrows}
\usetikzlibrary{shapes.multipart}
\tikzset{
state/.style={
       rectangle split,
       rectangle split parts=2,
       rectangle split part fill={red!30,blue!20},
       rounded corners,
       draw=black, very thick,
       minimum height=2em,
       text width=3cm,
       inner sep=2pt,
       text centered,
       }
}

\usepackage{xcolor}
\definecolor{Blue}{rgb}{0.0,0.0,1.0}
\definecolor{Red}{rgb}{1.0,0.0,0.0}
\newcommand{\mf}[1]{}

\def\abovestrut#1{\rule[0in]{0in}{#1}\ignorespaces}
\def\abovespace{\abovestrut{0.20in}}

\title{\nm: A Very Deep Hierarchy of Latent Variables for Generative Modeling}

\author{%
  Lars Maal{\o}e \\
  Corti\\
  Copenhagen\\
  Denmark \\
  \texttt{lm@corti.ai} \\
  \And
  Marco Fraccaro \\
  Unumed\\
  Copenhagen\\
  Denmark\\
  \texttt{mf@unumed.com}\\
  \And
  Valentin Li\'evin \& Ole Winther\\
  Technical University of Denmark\\
  Copenhagen\\
  Denmark\\
  \texttt{\{valv,olwi\}@dtu.dk}\\
}

\begin{document}

\maketitle

\begin{abstract} 
With the introduction of the variational autoencoder (VAE), probabilistic latent variable models have received renewed attention as powerful generative models. However, their performance in terms of test likelihood and quality of generated samples has been surpassed by autoregressive models without stochastic units. Furthermore, flow-based models have recently been shown to be an attractive alternative that scales well to high-dimensional data. In this paper we close the performance gap by constructing VAE models that can effectively utilize a deep hierarchy of stochastic variables and model complex covariance structures. We introduce the Bidirectional-Inference Variational Autoencoder (BIVA), characterized by a skip-connected generative model and an inference network formed by a bidirectional stochastic inference path. We show that BIVA reaches state-of-the-art test likelihoods, generates sharp and coherent natural images, and uses the hierarchy of latent variables to capture different aspects of the data distribution. We observe that BIVA, in contrast to recent results, can be used for anomaly detection. We attribute this to the hierarchy of latent variables which is able to extract high-level semantic features. Finally, we extend BIVA to semi-supervised classification tasks and show that it performs comparably to state-of-the-art results by generative adversarial networks.
\end{abstract}

\section{Introduction}

One of the key aspirations in recent machine learning research is to build models that \textit{understand the world} \citep{Kingma13, Rezende14,Goodfellow2014,Oord2015}. Generative models are providing the means to learn from a plethora of unlabeled data in order to model a complex data distribution, e.g. natural images, text, and audio. These models are evaluated by their ability to \textit{generate} data that is similar to the input data distribution from which they were trained on. The range of applications that come with generative models are vast, where audio synthesis \citep{Oord16b} and semi-supervised classification \citep{Rasmus15,Maaloe2016,Salimans2016} are examples hereof. Generative models can be broadly divided into explicit and implicit density models. The generative adversarial network (GAN) \citep{Goodfellow2014} is an example of an implicit model, since it is not possible to procure a likelihood estimation from this model framework. The focus of this research is instead within explicit density models, for which a tractable or approximate likelihood estimation can be performed.

The three main classes of powerful explicit density models are autoregressive models \citep{Larochelle11,Oord2015}, flow-based models \citep{Dinh14,Dinh16,Kingma18,Ho19}, and probabilistic latent variable models \citep{Kingma13,Rezende14,Mnih14}. In recent years autoregressive models, such as the PixelRNN and the PixelCNN \citep{Oord2015,Salimans17}, have achieved superior likelihood performance and flow-based models have proven efficacy on large-scale natural image generation tasks \citep{Kingma18}. However, in the autoregressive models, the runtime performance of generation is scaling poorly with the complexity of the input distribution. The flow-based models do not possess this restriction and do indeed generate visually compelling natural images when sampling close to the mode of the distribution. However, generation from the actual learned distribution is still not outperforming autoregressive models \citep{Kingma18, Ho19}.

Probabilistic latent variable models such as the variational auto-encoder (VAE) \citep{Kingma13,Rezende14} possess intriguing properties that are different from the other classes of explicit density models. They are characterized by a posterior distribution over the latent variables of the model, derived from Bayes' theorem, which is typically intractable and needs to be approximated.
This distribution most commonly lies on a low-dimensional manifold that can provide insights into the internal representation of the data \citep{Bengio2013a}. However, the latent variable models have largely been disregarded as powerful generative models due to \textit{blurry} generations and poor likelihood performances on natural image tasks. \citep{Larsen16,Dosovitskiy2016}, amongst others, attribute this tendency to the usage of a similarity metric in pixel space. Contrarily, we attribute it to the lack of overall model expressiveness for accurately modeling complex input distributions, as discussed in \citep{ZhaoSE17, rezende2018taming}.

There has been much research into explicitly defining and learning more expressive latent variable models. Here, the complementary research into learning a covariance structure through a framework of normalizing flows \citep{Rezende2015,Tomczak16,Kingma2016} and the stacking of a hierarchy of latent variables \citep{Burda15,Ranganath2015,Maaloe2016,Sonderby2016} have shown promising results. However, despite significant improvements, the reported performance of these models has still been inferior to their autoregressive counterparts. This has spawned a new class of explicit density models that adds an autoregressive component to the generative process of a latent variable model \citep{Gulrajani2016,Chen2017}. In this combination of model paradigms, the latent variables can be viewed as merely a  \textit{lossy} representation of the input data and the model still suffers from the same issues as autoregressive models.

\paragraph{Contributions. }In this research we argue that latent variable models that are defined in a sufficiently expressive way can compete with autoregressive and flow-based models in terms of test log-likelihood and quality of the generated samples.
We introduce the \name\ (\nm), a model formed by a deep hierarchy of stochastic variables that uses skip-connections to enhance the flow of information and avoid inactive units.  
To define a flexible posterior approximation, we construct a bidirectional inference network using stochastic variables in a bottom-up and a top-down inference path. The inference model is reminiscent to the stochastic top-down path introduced in the Ladder VAE \cite{Sonderby2016} and IAF VAE \cite{Sonderby2016} with the addition that the bottom-up pass is now also stochastic and there are no autoregressive components.
We perform an in-depth analysis of \nm and show
\textbf{(i)} an ablation study that analyses the contributions of the individual novel components,
\textbf{(ii)} that the model is able to improve on state-of-the-art results on benchmark image datasets, 
\textbf{(iii)} that a small extension of the model can be used for semi-supervised classification and performs comparably to current state-of-the-art models,  and 
\textbf{(iv)}  that the model, contrarily to other state-of-the-art explicit density models \citep{Nalisnick18}, can be utilized for anomaly detection on complex data distributions.

\section{Variational Autoencoders}
 The VAE is a generative model parameterized by a neural network $\theta$ and is defined by an observed variable $x$ that depends on a hierarchy of stochastic latent variables $\mathbf{z}=z_1, ..., z_L$ so that: $p_{\theta}(x, \mathbf{z}) = p_{\theta}(x|z_1) p_{\theta}(z_L) \prod_{i=1}^{L-1} p_{\theta}(z_i|z_{i+1})$. The posterior distribution over the latent variables of a VAE is commonly analytically intractable, and is approximated with a variational distribution
which is factorized with a bottom-up structure, $q_{\phi}(\mathbf{z}|x) = q_{\phi}(z_1|x)\prod_{i=1}^{L-1} q_{\phi}(z_{i+1}|z_i)$, 
so that each latent variable is conditioned on the variable below in the hierarchy. 
The parameters $\theta$ and $\phi$ can be optimized by maximizing the  \textit{evidence lower bound} (ELBO) 
\begin{align}
    \log p_{\theta}(x) \ge \mathbb{E}_{q_{\phi}(\mathbf{z}|x)}\left[\log \frac{p_{\theta}(x,\mathbf{z})}{q_{\phi}(\mathbf{z}|x)} \right] \equiv \mathcal{L}(\theta, \phi)\ . \label{eq:elbo}
\end{align}
A detailed introduction on VAEs can be found in appendix \ref{sec:vae_app} in the supplementary material. While a deep hierarchy of latent stochastic variables will result in a more expressive model, in practice the top stochastic latent variables of standard VAEs have a tendency to \textit{collapse} into the prior.
The Ladder VAE (LVAE) \citep{Sonderby2016} is amongst the first attempts towards VAEs that can effectively leverage multiple layers of stochastic variables. This is achieved by parameterizing the variational approximation with a \textit{bottom-up} deterministic path followed by a \textit{top-down} inference path that shares parameters with the top-down structure of the generative model: $q_{\phi,\theta}(\mathbf{z}|x) = q_{\phi}(z_L|x)\prod_{i=1}^{L-1} q_{\phi,\theta}(z_i|z_{i+1},x)$.
See Appendix \ref{sec:vae_app} for a graphical representation of the LVAE inference network.
Thanks to the bottom-up path, all the latent variables in the hierarchy have a deterministic dependency on the observed variable $x$, which allows data-dependent information to skip all the stochastic variables lower in the hierarchy (Figure \ref{fig:lvae_inf_skip} in Appendix \ref{sec:vae_app}).
The stochastic latent variables that are higher in the hierarchy will therefore receive less noisy inputs, and will be empirically less likely to collapse.
Despite the improvements obtained thanks to the more flexible inference network, in practice LVAEs with a very deep hierarchy of stochastic latent variables will still experience variable collapse. In the next section we will introduce the \name, that manages to avoid these issues by extending the LVAE in 2 ways: (i) adding a deterministic top-down path in the generative model and (ii) defining a factorization of the latent variables $z_i$ at each level of the hierarchy that allows to construct a bottom-up \textit{stochastic} inference path.

\section{\name}

\begin{figure}[!t]
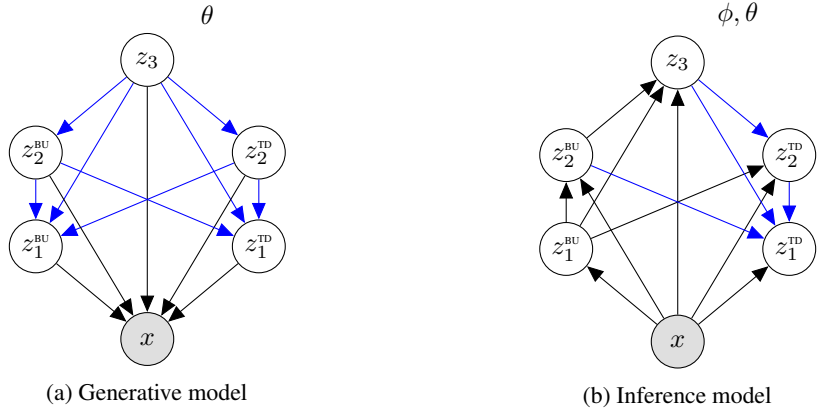

\begin{subfigure}{0.5\textwidth}
\centering	  
\def\col{blue}
      \tikz{
     \node[obs] (x) {$x$};%

     \node[above=.75cm of x] (d1){}; %
     \node[latent,right=1cm of d1] (z1td){$z_1^{\scalebox{0.4}{TD}}$}; %
     \node[latent,left=1cm of d1](z1bu){$z_1^{\scalebox{0.4}{BU}}$}; %
     
     \node[above=1.cm of d1] (d2){}; %
     \node[latent,right=1cm of d2] (z2td){$z_2^{\scalebox{0.4}{TD}}$}; %
     \node[latent,left=1cm of d2](z2bu){$z_2^{\scalebox{0.4}{BU}}$}; %
     \node[latent,above=.75cm of d2] (z3td){$z_3$}; %
     
     \node[above=of z3td, xshift=.8cm, yshift=-1.cm] (theta) {$\theta$}; %
     
     \edge[\col]{z3td}{z2bu}
     \edge[\col]{z3td}{z2td}
     \edge[\col]{z3td}{z1bu}
     \edge[\col]{z3td}{z1td}
     \edge[]{z3td}{x}

     \edge[\col]{z2td}{z1bu}
     \edge[\col]{z2td}{z1td}
     \edge[]{z2td}{x}

     \edge[\col]{z2bu}{z1bu}
     \edge[\col]{z2bu}{z1td}
     \edge[]{z2bu}{x}

     \edge[]{z1td}{x}
     \edge[]{z1bu}{x}

     }
     \caption{Generative model}\label{fig:lvae++_gen_graph}
\end{subfigure}
\begin{subfigure}{0.5\textwidth}
\centering	  
\def\col{blue}
      \tikz{
     \node[obs] (x) {$x$};%

     \node[above=.75cm of x] (d1){}; %
     \node[latent,right=1cm of d1] (z1td){$z_1^{\scalebox{0.4}{TD}}$}; %
     \node[latent,left=1cm of d1](z1bu){$z_1^{\scalebox{0.4}{BU}}$}; %
     
     \node[above=1.cm of d1] (d2){}; %
     \node[latent,right=1cm of d2] (z2td){$z_2^{\scalebox{0.4}{TD}}$}; %
     \node[latent,left=1cm of d2](z2bu){$z_2^{\scalebox{0.4}{BU}}$}; %
     \node[latent,above=.75cm of d2] (z3td){$z_3$}; %
     
     \node[above=of z3td, xshift=.8cm, yshift=-1.cm] (phi) {$\phi, \theta$}; %

     \edge[]{x}{z1bu}

     \edge[]{x}{z2bu}
     \edge[]{z1bu}{z2bu}

     \edge[]{x}{z3td}
     \edge[]{z2bu}{z3td}
     \edge[]{z1bu}{z3td}

     \edge[]{x}{z2td}
     \edge[]{z1bu}{z2td}
     \edge[\col]{z3td}{z2td}
     
     \edge[]{x}{z1td}
     \edge[\col]{z2bu}{z1td}
     \edge[\col]{z3td}{z1td}
     \edge[\col]{z2td}{z1td}

     }
     \caption{Inference model}\label{fig:lvae++_inf_graph}
\end{subfigure}
\caption{A $L=3$ layered \nm with (a) the generative model and (b) inference model. {\color{blue} Blue} arrows indicate that the deterministic parameters are shared between the inference and generative models. See Appendix \ref{app:model_description} for a detailed explanation and a graphical model that includes the deterministic variables.}\label{fig:lvae++}
\vspace{-.5cm}
\end{figure}

In this section, we will first describe the architecture of the \name\ (Figure \ref{fig:lvae++}), and then provide the motivation behind the main ideas of the model as well as some intuitions on the role of each of its novel components. Finally, we will show how this model can be used for a novel approach to detecting anomalous data.

\subsection{Model architecture}
\paragraph{Generative model.} In \nm, at each layer $1,...,L-1$ of the hierarchy we split the latent variable in two components, $z_i = (z_{i}^{\scalebox{0.6}{BU}}, z_{i}^{\scalebox{0.6}{TD}})$, which belong to a bottom-up (BU) and top-down (TD) inference path, respectively. More details on this will be given when introducing the inference network. 
The generative model of \nm is illustrated in Figure \ref{fig:lvae++_gen_graph}.
We introduce a deterministic top-down path $d_{L-1}, \dots, d_1$ that is parameterized with neural networks and receives as input at each layer $i$ of the hierarchy the latent variable $z_{i+1}$. In the case of a convolutional model, this is done by concatenating $(z_{i+1}^{\scalebox{0.6}{BU}}$, $z_{i+1}^{\scalebox{0.6}{TD}})$ and $d_{i+1}$ along the features' dimension. $d_i$ can therefore be seen as a deterministic variable that summarizes all the relevant information coming from the stochastic variables higher in the hierarchy, $z_{>i}$.  The latent variables $z_i^{\scalebox{.6}{BU}}$ and $z_i^{\scalebox{.6}{TD}}$ are conditioned on all the information in the higher layers, and are conditionally independent given $z_{>i}$.
The joint distribution of the model is then given by:
\begin{align*}
p_{\theta}(x,\mathbf{z}) = 
p_{\theta}(x|\mathbf{z}) p_{\theta}(z_L) \prod_{i=1}^{L-1}
p_{\theta}(z_i^{\scalebox{.6}{BU}}|z_{>i})p_{\theta}(z_i^{\scalebox{.6}{TD}}|z_{>i}) \ , 
\end{align*}
where $\theta$ are the parameters of the generative model.
The likelihood of the model $p_{\theta}(x|\mathbf{z})$ directly depends on $z_1$, and depends on $z_{>1}$ through the deterministic top-down path.
Each stochastic latent variable $1,...,L$ is parameterized by a Gaussian distribution with diagonal covariance, with one neural network $\mu(\cdot)$ for the mean and another neural network $\sigma(\cdot)$ for the variance.
Since the $z_{i+1}^{\scalebox{0.6}{BU}}$ and $z_{i+1}^{\scalebox{0.6}{TD}}$ variables are on the same level in the generative model and of the same dimensionality, we share all the deterministic parameters going to the layer below. See Appendix \ref{app:model_description} for details.

\paragraph{Bidirectional inference network.}
Due to the non-linearities in the neural networks that parameterize the generative model, the exact posterior distribution $p_\theta(\mathbf{z}|x)$ is intractable and needs to be approximated. As for VAEs, we therefore define a variational distribution, $q_{\phi}(\mathbf{z}|x)$, that needs to be flexible enough to approximate the true posterior distribution, as closely as possible.
We define a bottom-up (BU) and a top-down (TD) inference path, which are computed sequentially when constructing the posterior approximation for each data point $x$, see Figure \ref{fig:lvae++_inf_graph}. The variational distribution over the BU latent variables depends on the data $x$ and on all BU variables lower in the hierarchy, i.e. $q_{\phi}(z_i^{\scalebox{.6}{BU}}|x,z_{<i}^{\scalebox{.6}{BU}})$, where $\phi$ denotes all the parameters of the BU path. $z_i^{\scalebox{.6}{BU}}$ has a direct dependency only on the BU variable below, $z_{i-1}^{\scalebox{.6}{BU}}$. The dependency on $z_{<i-1}^{\scalebox{.6}{BU}}$ is achieved, similarly to the generative model, through a deterministic bottom-up path $\widetilde{d}_{1}, \dots, \widetilde{d}_{L-1}$.

The TD variables depend on the data and the BU variables lower in the hierarchy through the BU inference path, but also on all variables above in the hierarchy through the TD inference path, see Figure \ref{fig:lvae++_inf_graph}. The variational approximation over the TD variables is thereby $q_{\phi,\theta}(z_i^{\scalebox{.6}{TD}}|x,z_{<i}^{\scalebox{.6}{BU}},z_{>i}^{\scalebox{.6}{BU}}, z_{>i}^{\scalebox{.6}{TD}})$. Importantly, all the parameters of the TD path are shared with the generative model, and are therefore denoted as $\theta$. The overall inference network can be factorized as follows:
\begin{align*}
q_{\phi}(\mathbf{z}|x) = q_{\phi}(z_L|x,z_{<L}^{\scalebox{.6}{BU}}) \prod_{i=1}^{L-1}  q_{\phi}(z_i^{\scalebox{.6}{BU}}|x,z_{<i}^{\scalebox{.6}{BU}}) q_{\phi,\theta}(z_i^{\scalebox{.6}{TD}}|x,z_{<i}^{\scalebox{.6}{BU}},z_{>i}^{\scalebox{.6}{BU}}, z_{>i}^{\scalebox{.6}{TD}})\ ,
\end{align*}
where the variational distributions over the BU and TD latent variables are Gaussians whose mean and diagonal covariance are parameterized with neural networks that take as input the concatenation over the feature dimension of the conditioning variables.  
Training of \nm is performed, as for VAEs, by maximizing the ELBO in eq. \eqref{eq:elbo} with stochastic backpropagation and the reparameterization trick.

\subsection{Motivation}
\nm can be seen as an extension of the LVAE in which we (i) add a deterministic top-down path and (ii) apply a bidirectional inference network. We will now provide the motivation and some intuitions on the role of these two novel components, that will then be empirically validated with the ablation study of Section \ref{sec:ablation}. 

\paragraph{Deterministic top-down path.}
Skip-connections represent one of the simplest yet most powerful advancements of deep learning in recent years.  They allow constructing very deep neural networks, by better propagating the information throughout the model and reducing the issue of vanishing gradients. Skip connections form for example the backbone of deep neural networks such as ResNets \citep{He2015}, which have shown impressive performances on a wide range of classification tasks.
Our goal in this paper is to build very deep latent variable models that are able to learn an expressive latent hierarchical representation of the data. In our experiments, we however found that the LVAE still had difficulties in activating the top latent variables for deeper hierarchies. To limit this issue, we add skip connections among the latent variables in the generative model by adding the deterministic top-down path, that makes each variable depend on all the variables above in the hierarchy (see Figure \ref{fig:lvae++_gen_graph} for a graphical representation). This allows a better flow of information in the model and thereby avoids the collapse of latent variables.
A related idea was recently proposed by \citep{Dieng18}, that add skip connections among the neural network layers parameterizing a shallow VAE with a single latent variable.

\paragraph{Bidirectional inference.}
The inspiration for the bidirectional inference network of \nm comes from the work on Auxiliary VAEs (AVAE) by \citep{Ranganath2015,Maaloe2016}.
An AVAE can be viewed as a shallow VAE with a single latent variable $z$ and an auxiliary variable $a$ that increases the expressiveness of the variational approximation $q_{\phi}(z|x) = \int q_{\phi}(z|a,x) q_{\phi}(a|x)  \mathrm{d} a$. By making the inference network $q_{\phi}(z|a,x)$ depend on the stochastic variable $a$, the AVAE adds covariance structure to the posterior approximation over the stochastic unit $z$, since it no longer factorizes over its components $z^{(k)}$, i.e. $q_{\phi}(z|x) \neq \prod_{k} q_{\phi}(z^{(k)}|x)$.
As discussed in the following, by factorizing the latent variables at each level of the hierarchy of \nm we are able to achieve similar results without introducing additional auxiliary variables in the model. To see this, we can focus for example on the highest latent variable $z_L$.
In \nm, the presence of the $z_{i}^{\scalebox{.6}{BU}}$ variables makes the bottom-up inference path \textit{stochastic}, as opposed to the deterministic BU path of the LVAE. 
While the conditional distribution $q_{\phi}(z_L|x,z_{<L}^{\scalebox{.6}{BU}})$ still factorizes over the components of $z_L$, due to the stochastic BU variables the marginal distribution over $z_L$ no longer factorizes, i.e.
$
q_\phi(z_L|x)=\int q_{\phi}(z_L|x,z_{<L}^{\scalebox{.6}{BU}})
q_{\phi}(z_{<L}^{\scalebox{.6}{BU}}|x) \mathrm{d} z_{<L}^{\scalebox{.6}{BU}} \neq \prod_{k=1}^K q(z_{L}^{(k)}|x) \ .
$
Therefore, the BU inference path enables the learning of a complex covariance structure in the higher TD stochastic latent variables, which is fundamental in the model to extract \textit{good} high-level semantic features from the data distribution. Notice that, in \nm, only $z_1^{\scalebox{.6}{BU}}$ will have a marginally factorizing inference network.

\subsection{Anomaly detection with \nm}
Anomaly detection is considered to be one of the most important applications of explicit density models. 
However, recent empirical results suggest that these models are not able to distinguish between two clearly distinctive data distributions  \citep{Nalisnick18}, as they can assign a higher likelihood to data points from a data distribution that is very different from the one the model was trained on. 
Based on a thorough study, \citep{Nalisnick18} states that the main issue is the fact that explicit density models tend to capture low-level statistics, as opposed to the high-level semantics that are preferable when doing anomaly detection.
We hypothesize that the latent representations in the higher layers of \nm 
can capture the high-level semantics of the data and that these can be used for improved anomaly detection.  

In the standard ELBO from eq.~\eqref{eq:elbo}, the main contribution to the expected log-likelihood term is coming from averaging over the variational distribution of the lower level latent variables. This will thus emphasize low-level statistics. So in order to perform anomaly detection with \nm we instead need to emphasize the contribution from the higher layers. We can achieve this with an alternative log-likelihood lower bound that partly replaces the inference network with the generative model. It will be a weaker bound than the ELBO, but it has the advantage that it explicitly uses the generative hierarchy of the stochastic variables.
In the following we define  
the hierarchy of stochastic latent variables 
as $\mathbf{z}=z_1, z_2, z_3, ..., z_L$ with $z_i =(z_i^{\scalebox{0.6}{BU}},z_i^{\scalebox{0.6}{TD}})$.
Instead of using the variational approximation $q_{\phi}(\mathbf{z}|x)$ over all stochastic variables in the model, we use the prior distribution for the first $k$ layers and the variational approximation for the others, i.e.~$p_\theta(z_{\leq k}|z_{>k})q_{\phi}(z_{>k}|x)$. 
The new ELBO becomes:
\begin{align}
    \mathcal{L}^{>k} = \mathbb{E}_{p_\theta(z_{\leq k}|z_{>k})q_{\phi}(z_{>k}|x)}\left[\log \frac{p_{\theta}(x|\mathbf{z})p_{\theta}(z_{>k})}{q_{\phi}(z_{>k}|x)} \right]\ . \label{eq:anomaly}
\end{align}
$\mathcal{L}^{>0}=\mathcal{L}$ is the ELBO in eq. \eqref{eq:elbo}. As for the ELBO, we approximate the computation of $\mathcal{L}^{>k}$ with Monte Carlo integration. 
Sampling from $p_\theta(z_{\leq k}|z_{>k})q_{\phi}(z_{>k}|x)$ can be easily performed by obtaining samples $\widehat{z}_{>k}$ from the inference network, that are then used to sample $\widehat{z}_{\leq k}$ from the conditional prior $p_\theta(z_{\leq k}|\widehat{z}_{>k})$.

Due to the sampling from the prior, eq. \eqref{eq:anomaly} will generally return a worse likelihood approximation than the ELBO. Despite this, $\mathcal{L}^{>k}$ with higher values of $k$ represents a useful metric for anomaly detection. 
By only sampling the top $L-k$ variables from the variational approximation, in fact, we are forcing the model to only rely on the high-level semantics encoded in the highest variables of the hierarchy when evaluating this metric, and not on the low-level statistics encoded in the lower variables. 

\section{Experiments}\label{sec:experiments}

\begin{figure}
  \begin{minipage}[c]{0.56\textwidth}
\begin{subfigure}{0.32\textwidth}
\centering
\includegraphics[width=1.\textwidth]{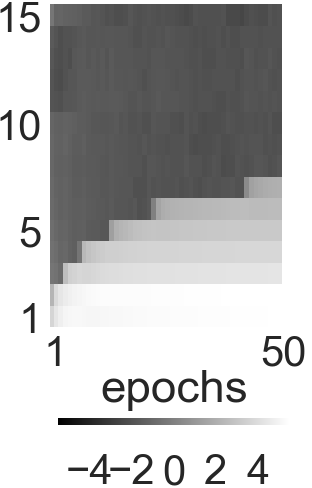}
\caption{LVAE $L=15$}
\end{subfigure}
\hfill
\begin{subfigure}{0.32\textwidth}
\centering
\includegraphics[width=1.\textwidth]{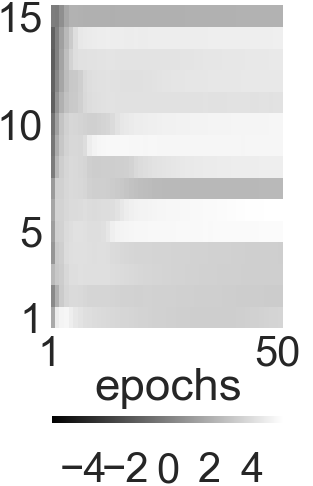}
\caption{LVAE+ $L=15$}
\end{subfigure}
\hfill
\begin{subfigure}{0.32\textwidth}
\centering
\includegraphics[width=1.\textwidth]{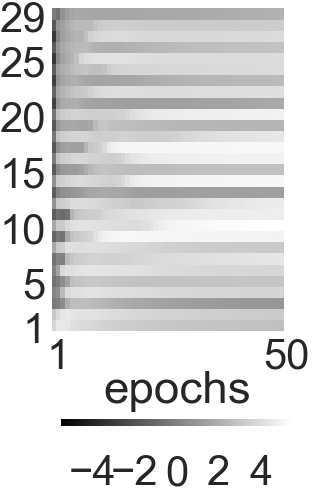}
\caption{\nm $L=15$}
\end{subfigure}
\end{minipage}\hfill
  \begin{minipage}[c]{0.41\textwidth}
\caption{The $\log KL(q||p)$ for each stochastic latent variable as a function of the training epochs on CIFAR-10. (a) is a $L=N=15$ stochastic latent layer LVAE with no skip-connections and no bottom-up inference. (b) is a $L=N=15$ LVAE+ with skip-connections and no bottom-up inference. (c) is a $L=15$ stochastic latent layer ($N=29$ latent variables) \nm for which $1,2,...,N$ denotes the stochastic latent variables following the order $z_1^{\protect\scalebox{0.4}{BU}}, z_1^{\protect\scalebox{0.4}{TD}}, z_2^{\protect\scalebox{0.4}{BU}}, z_2^{\protect\scalebox{0.4}{TD}}, ..., z_L$.}
\label{fig:latent_variables}
  \end{minipage}
  \vspace*{-0.5cm}
\end{figure}

\begin{figure*}[!t]
\centering
\includegraphics[width=1.\textwidth]{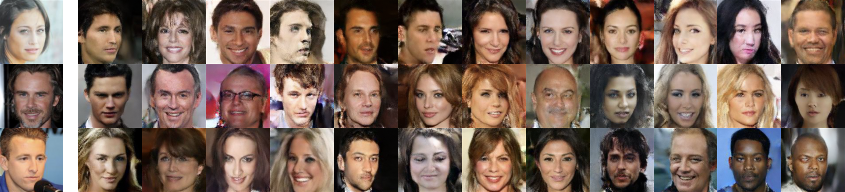}
\caption{(left) images from the CelebA dataset preprocessed to 64x64 following \citep{Larsen16}. (right) $\mathcal{N}(0,I)$ generations of \nm with $L=20$ layers that achieves a $\mathcal{L}_1=2.48$ bits/dim on the test set.}
\label{fig:celeb_gen}
	\vspace*{-0.4cm}
\end{figure*}

\nm is empirically evaluated by (i) an ablation study analyzing each novel component, (ii) likelihood and semi-supervised classification results on binary images, (iii) likelihood results on natural images, and (iv) an analysis of anomaly detection in complex data distributions. 
We employ a \textit{free bits} strategy with $\lambda=2$ \citep{Kingma2016} for all experiments to avoid latent variable collapse during the initial training epochs.
Trained models are reported with 1 importance weighted sample, $\mathcal{L}_1$, and 1000 importance weighted samples, $\mathcal{L}_{1e3}$ \citep{Burda15a}. We evaluate the natural image experiments by bits per dimension (bits/dim), $\mathcal{L}/(hwc\log(2))$, where $h$, $w$, $c$ denote the height, width, and channels respectively.
For a detailed description of the experimental setup see Appendix \ref{app:experimental_setup} and the source code\footnote{Source code (Tensorflow): https://github.com/larsmaaloee/BIVA.}\footnote{Source code (PyTorch): https://github.com/vlievin/biva-pytorch.}.
In Appendix \ref{app:2d_density} we test \nm  on complex 2d densities, while Appendix \ref{app:text_modeling} presents initial results for the model on text.

\subsection{Ablation Study}\label{sec:ablation}
\nm can be viewed as an extension of the LVAE from \citep{Sonderby2016} where we add (i) extra dependencies in the generative model ($p_\theta(x|z_1)\to p_\theta(x|\mathbf{z})$ and $p_\theta(z_i|z_{i+1})\to p_\theta(z_i|z_{>i})$) through the skip connections obtained with the deterministic top-down path and (ii) a bottom-up (BU) path of stochastic latent variables to the inference model. 
In order to evaluate the effects of each added component we define an LVAE with the exact same architecture as \nm, but without the BU variables and the deterministic top-down path. Next, we define the LVAE+, where we add to the LVAE's generative model the deterministic top-down path. It is therefore the same model as in Figure \ref{fig:lvae++} but without the BU variables. Finally, we investigate a LVAE+ model with $2L-1$ stochastic layers. This corresponds to the depth of the hierarchy of the \nm inference model $x\to z^{\scalebox{0.4}{BU}}_1\to \dots \to z^{\scalebox{0.4}{BU}}_{L-1}\to z_L \to z^{\scalebox{0.4}{TD}}_{L-1} \to \dots \to z^{\scalebox{0.4}{TD}}_{1}$. 
If this model is competitive with \nm then it is an indication that it is the depth that determines the performance. 
The ablation study is conducted on the CIFAR-10 dataset against the best reported \nm with $L=15$ layers (Section \ref{sec:natural_images}), which means $2L-1=29$ stochastic latent layers in the deep LVAE+.

\begin{wraptable}{r}{0.45\textwidth}
\vspace*{-0.3cm}
{\caption{A comparison of the LVAE with no skip-connections and no bottom-up inference, the LVAE+ with skip-connections and no bottom-up inference, and \nm. All models are trained on the CIFAR-10 dataset.}\label{table:ablation}}
\begin{small}
\begin{sc}
\vspace*{-0.2cm}
\begin{tabular}{l c c}
& Param. & bits/dim \\
\hline
LVAE L=15, $\mathcal{L}_1$ & 72.36M & $ \leq 3.60$ \\
LVAE+ L=15, $\mathcal{L}_1$ & 73.35M & $\leq 3.41$ \\
LVAE+ L=29, $\mathcal{L}_1$ & 119.71M & $ \leq 3.45$\\
\nm L=15, $\mathcal{L}_1$ & 102.95M & $ \leq 3.12$\\
\hline
\end{tabular}%
\end{sc}
\end{small}
\vspace*{-0.4cm}
\end{wraptable} 
Table \ref{table:ablation} presents a comparison of the different model architectures. The positive effect of adding the skip connections in the generative models can be evaluated from the difference between the LVAE $L=15$ and LVAE+ $L=15$ results, for which there is close to a 0.2 bits/dim difference in the ELBO. 
Thanks to the more expressive posterior approximation obtained using its bidirectional inference network, \nm improves the ELBO significantly w.r.t the LVAE+, by more than 0.3 bits/dim.
Notice that a deeper hierarchy of stochastic latent variables in the LVAE+ will not necessarily provide a better likelihood performance, since the LVAE+ $L=29$ performs worse than the LVAE+ $L=15$ despite having significantly more parameters. In Figure \ref{fig:latent_variables} we plot for LVAE, LVAE+ and \nm the KL divergence between the variational approximation over each latent variable and its prior distribution, $KL(q||p)$. This KL divergence is 0 when the two distributions match, in which case we say that the variable has collapsed, since its posterior approximation is not using any data-dependent information. We can see that while the LVAE is only able to utilize its lowest 7 stochastic variables, all variables in both LVAE+ and \nm are active. 
We attribute this tendency to the deterministic top-down path that is present in both models, which creates skip-connections between all latent variables that allow to better propagate the information throughout the model.

\subsection{Binary Images}
\begin{table}[t]
\begin{minipage}{0.48\textwidth}
\begin{small}
\begin{sc}
{\caption{Test log-likelihood on statically binarized MNIST for different number of importance weighted samples. The finetuned models are trained for an additional number of epochs with no \textit{free bits}, $\lambda=0$. For testing resiliency we trained 4 models and evaluated the standard deviations to be $\pm 0.031$ for $\mathcal{L}_1$.}\label{table:bin_mnist}}
\begin{tabular}{l c}
 & $-\log p(x)$ \\
\hline
\textit{With autoregressive components} \\
PixelCNN {\scriptsize\cite{Oord2015}} & $=81.30$\\
DRAW {\scriptsize\cite{Gregor15}} & $< 80.97$\\
IAFVAE {\scriptsize\cite{Kingma2016}} & $\leq 79.88$\\
PixelVAE {\scriptsize\cite{Gulrajani2016}} & $\leq 79.66$\\
PixelRNN {\scriptsize\cite{Oord2015}} & $=79.20$\\
VLAE {\scriptsize\cite{Chen2017}} & $\leq 79.03$\\
 \hline
 \textit{Without autoregressive components} \\
 Discrete VAE {\scriptsize\cite{Rolfe2017}} & $\leq 81.01$\\
\abovespace
\textbf{\nm}, $\mathcal{L}_1$ & $\leq 81.20$ \\
\textbf{\nm}, $\mathcal{L}_{1e3}$ & $\leq 78.67$ \\
\textbf{\nm} finetuned, $\mathcal{L}_1$ & $\leq 80.47$ \\
\textbf{\nm} finetuned, $\mathcal{L}_{1e3}$ & $\leq 78.59$ \\
\hline
\end{tabular}%
\end{sc}
\end{small}
\end{minipage}
\hfill
\begin{minipage}{0.49\textwidth}
\vspace*{-2.4cm}
\caption{Semi-supervised test error for \nm on MNIST for 100 randomly chosen and evenly distributed labelled samples.}\label{table:semisupervised}
\vspace{-0.6cm}
\begin{center}
\begin{small}
\begin{sc}
\begin{tabular}{ll}
\abovespace
   &  Error \%\\
\hline
\abovespace 
M1+M2 {\scriptsize\citep{Kingma14}} & $3.33$\% ($\pm 0.14$) \\
VAT {\scriptsize\citep{Miyato15}} & $2.12$\% \\
CatGAN {\scriptsize\citep{Springenberg2015}} & $1.91$\% ($\pm 0.10$) \\
SDGM {\scriptsize\citep{Maaloe2016}} & $1.32$\% ($\pm 0.07$) \\
LadderNet {\scriptsize\citep{Rasmus15}} & $1.06$\% ($\pm 0.37$)\\ 
ADGM {\scriptsize\citep{Maaloe2016}} & $0.96$\% ($\pm 0.02$) \\
ImpGAN {\scriptsize\citep{Salimans2016}} & $0.93$\% ($\pm 0.07$) \\
TripleGAN {\scriptsize\citep{Li17}} & $0.91$\% ($\pm 0.58$)\\
SSLGAN {\scriptsize\citep{Dai17}} & $0.80$\% ($\pm 0.10$) \\
\abovespace
\textbf{\nm} & $0.83$\% ($\pm 0.02$)\\
\hline
\end{tabular}
\end{sc}
\end{small}
\end{center}
        \end{minipage}

    	\begin{minipage}{0.48\linewidth}
		\centering
		\includegraphics[width=.8\textwidth]{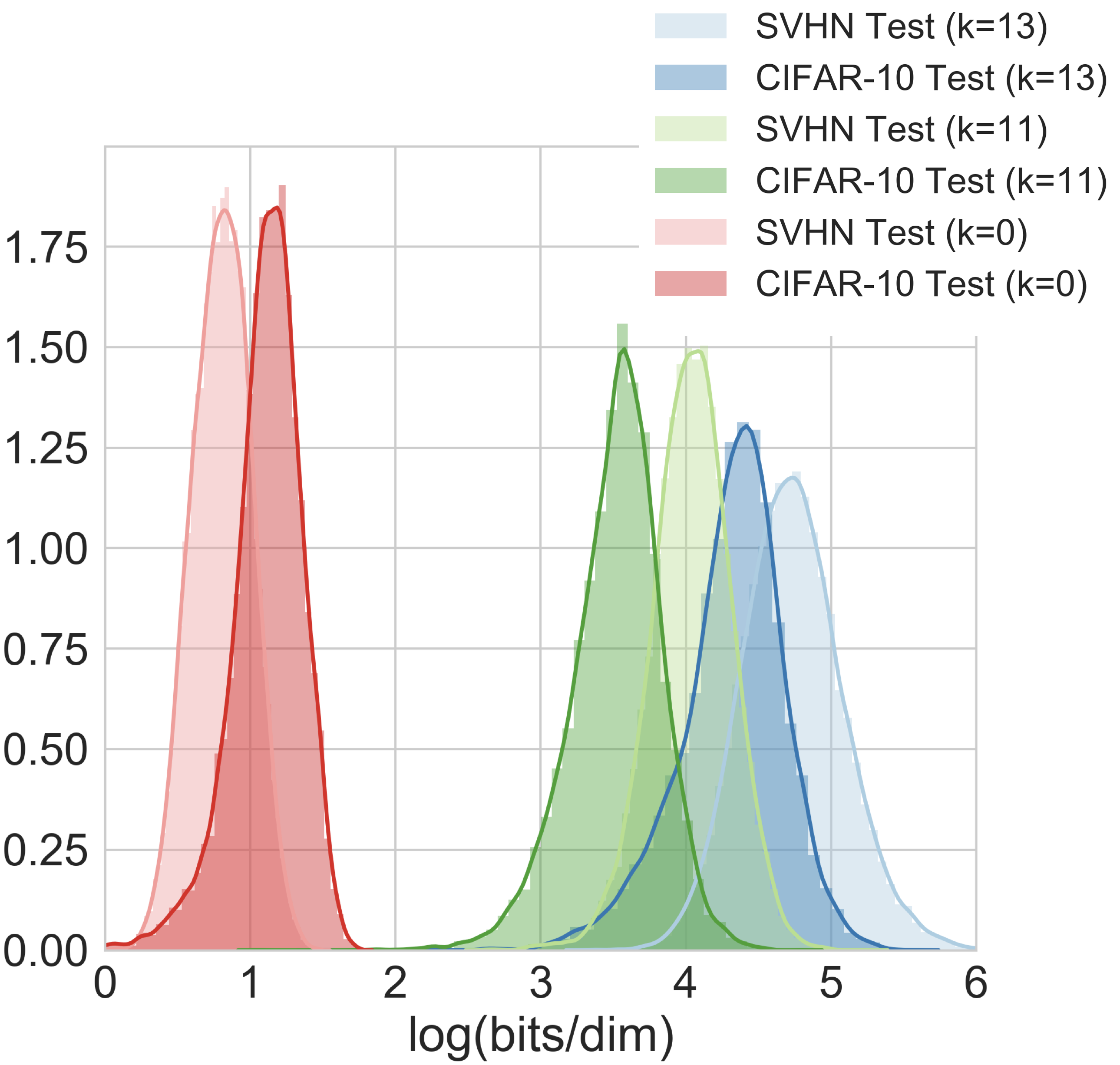}
		\captionof{figure}{Histograms and kernel density estimation of the $\mathcal{L}^{>k}$ for $k=13,11,0$ evaluated in bits/dim by a model trained on the CIFAR-10 train dataset and evaluated on the CIFAR-10 and the SVHN test set.}
		\label{fig:anomaly}
	\end{minipage}
	\hfill
	\begin{minipage}{0.48\linewidth}
		    \vspace*{-1.8cm}
\begin{small}
\vspace{-0.8cm}
\begin{sc}
\caption{Test log-likelihood on CIFAR-10
for different number of importance weighted samples. We evaluated two different \nm with various number of layers ($L$). For testing resiliency we trained 3 models and evaluated the standard deviations to be $\pm 0.013$ for $\mathcal{L}_1$ and $L=15$.
}\label{table:cifar}
		    \vspace*{-0.1cm}
\begin{tabular}{l c}
 & bits/dim \\
\hline
\textit{With autoregressive components} \\
ConvDRAW {\scriptsize\cite{Gregor16}} & $<3.58$\\
IAFVAE $\mathcal{L}_1$ {\scriptsize\cite{Kingma2016}} & $\leq 3.15$\\
IAFVAE $\mathcal{L}_{1e3}$ {\scriptsize\cite{Kingma2016}} & $\leq 3.12$\\
GatedPixelCNN {\scriptsize\cite{Oord16}} & $= 3.03$\\
PixelRNN {\scriptsize\cite{Oord2015}} & $= 3.00$\\
VLAE {\scriptsize\cite{Chen2017}} & $\leq 2.95$\\
PixelCNN++ {\scriptsize\cite{Salimans17}} & $= 2.92$\\
 \hline
 \textit{Without autoregressive components} \\
NICE {\scriptsize\cite{Dinh14}} & $=4.48$\\
DeepGMMs {\scriptsize\cite{Oord14}} & $=4.00$\\
RealNVP {\scriptsize\cite{Dinh16}} & $=3.49$\\
DiscreteVAE++ {\scriptsize\cite{Vahdat18}} & $\leq 3.38$\\
GLOW {\scriptsize\cite{Kingma18}} & $=3.35$\\
Flow++ {\scriptsize\cite{Ho19}} & $=3.08$\\
 \abovespace
\textbf{\nm} L=10, $\mathcal{L}_1$ & $\leq 3.17$ \\
\textbf{\nm} L=15, $\mathcal{L}_1$ & $\leq 3.12$ \\
\textbf{\nm} L=15, $\mathcal{L}_{1e3}$ & $\leq 3.08$ \\
\hline
\end{tabular}%
\end{sc}
\end{small}
	\end{minipage}
	\vspace*{-0.5cm}
\end{table}

We evaluate \nm $L=6$ in terms of test log-likelihood on statically binarized MNIST \citep{Salakhutdinov08}, dynamically binarized MNIST \citep{LeCun98} and dynamically binarized OMNIGLOT \citep{Lake2013}. The model parameterization and optimization parameters have been kept identical for all binary image experiments (see Appendix \ref{app:experimental_setup}). For each experiment on binary image datasets, we \textit{finetune} each model by setting the free bits to $\lambda=0$ until convergence in order to test the tightness of the $\mathcal{L}_1$ ELBO.

To the best of our knowledge, \nm achieves state-of-the-art results on statically binarized MNIST,  outperforming other latent variable models, autoregressive models, and flow-based models (see Table \ref{table:bin_mnist}). Finetuning the model with $\lambda=0$ improves the $\mathcal{L}_1$ ELBO significantly and achieves slightly better performance for the 1000 importance weighted samples. For dynamically binarized MNIST and OMNIGLOT, \nm achieves similar improvements with $\mathcal{L}_{\text{1e3}}=78.41$ (state-of-the-art) and $\mathcal{L}_{\text{1e3}}=91.34$ respectively, see Tables \ref{table:dyn_mnist} and \ref{table:omniglot} in Appendix \ref{app:additional_results}.

\paragraph{Semi-supervised learning.}
\nm can be easily extended for semi-supervised classification by adding a categorical variable $y$ to represent the class, as done in \citep{Kingma14}. We add a classification model $q_{\phi}(y|x,z_{<L}^{\scalebox{.6}{BU}})$ to the inference network, and a class-conditional distribution $p_{\theta}(x|\mathbf{z},y)$ to the generative model (see Appendix \ref{app:semi_supervised} for a detailed description). 
We train 5 different semi-supervised models on MNIST, each using a different set of just 100 randomly chosen and evenly distributed MNIST labels. Table \ref{table:semisupervised} presents the classification results on the test set (mean and standard deviation over the 5 runs), that shows that \nm achieves comparable performance to recent state-of-the-art results by generative adversarial networks.

\subsection{Natural Images}\label{sec:natural_images}

We trained and evaluated \nm $L=15$ on 32x32 CIFAR-10, 32x32 ImageNet \citep{Oord2015}, and another \nm $L=20$ on 64x64 CelebA \citep{Larsen16}. For the output decoding, we employ the discretized logistic mixture likelihood from \citep{Salimans17} (see Appendix \ref{app:experimental_setup} for more details). In Table \ref{table:cifar} we see that for the CIFAR-10 dataset \nm outperforms other state-of-the-art non-autoregressive models and performs slightly worse than state-of-the-art autoregressive models. For the 32x32 ImageNet dataset \nm achieves better performance than flow-based models, but the performance gap to the autoregressive models remains large (Table \ref{table:imagenet} in Appendix \ref{app:additional_results}). This may be due to the added complexity (more categories) of the 32x32 ImageNet dataset, requiring an even more flexible model. More research should be invested in defining an improved architecture for \nm that holds more parameters and thereby achieves better performances. 

Figure \ref{fig:celeb_gen} shows generated samples from the $\mathcal{N}(0,I)$ prior of a \nm $L=20$ trained on the CelebA dataset. From a visual inspection, the samples are far superior to previous natural image generations by latent variable models. We believe that previous claims stating that this type of model can only generate \textit{blurry} images should be disregarded \citep{Larsen16}. Rather the limited expressiveness/flexibility of previous models should be blamed. Additional samples from \nm can be found in Appendix \ref{app:additional_results}.

\subsection{Does \nm know what it doesn't know?}

We test the anomaly detection capabilities of \nm replicating the most challenging experiments of \citep{Nalisnick18}.
We train \nm $L=15$ on the CIFAR-10 dataset, and evaluate eq.~\eqref{eq:anomaly} for various values of $k$ on the CIFAR-10 test set, the SVHN dataset \citep{Netzer2011} and the CelebA dataset. The results can be found in Table \ref{table:anomaly} and Figure \ref{fig:anomaly}, and are reported in terms of bits per dimension (lower is better).
We see that for $k=0$, corresponding to the standard ELBO, \nm wrongly assigns lower values to data points from SVHN. This is in line with the results obtained with other explicit density models in \citep{Nalisnick18}, and shows that by using the standard ELBO the low-level image statistics prevail and the model is not able to correctly detect out-of-distribution samples. However, for higher values of $k$, the situation is reversed. We take this as an indication that \nm uses the high-level semantics inferred from the data to better differentiate between the CIFAR-10 and the SVHN/CelebA distributions.
We repeat the experiment training \nm $L=6$ on the FashionMNIST dataset (Table \ref{table:anomaly}), and testing on the FashionMNIST test set and the MNIST dataset. Unlike the flow-based models used in \citep{Nalisnick18}, \nm is able to learn a data distribution that can be used to detect anomalies with the standard ELBO (but also $k>0$).

\begin{table}[t]
  \begin{minipage}[c]{0.75\textwidth}
\begin{small}
\begin{sc}

\begin{tabular}{l l l l l}
& $\mathcal{L}^{>L-2}$ & $\mathcal{L}^{>L-4}$ & $\mathcal{L}^{>L-6}$ & $\mathcal{L}^{>0}$\\
\hline
\textit{Model trained on CIFAR-10:} \\
CIFAR-10 & 79.36 & 35.34 & 20.93 & 3.12\\
SVHN & 121.04 & 58.82 & 26.76 & 2.28 \\
\textit{Model trained on FashionMNIST:} \\
FashionMNIST & 228.38 & 107.07 & - & 94.05 \\
MNIST & 295.95 & 130.39 & - & 128.60 \\
\hline
\end{tabular}
\end{sc}
\end{small}
\end{minipage}\hfill
  \begin{minipage}[c]{0.25\textwidth}
\caption{The test $\mathcal{L}^{>k}$ for different values of $k$ and train/test dataset combinations evaluated in bits/dim for natural images and negative log-likelihood for binary images (lower is better).}\label{table:anomaly}
  \end{minipage}
  \vspace*{-0.6cm}
\end{table}

\section{Conclusion}
In this paper, we have introduced \nm, that significantly improves performances over previously introduced probabilistic latent variable models and flow-based models. \nm is able to generate natural images that are both sharp and coherent, to improve on semi-supervised classification benchmarks and, contrarily to other models, allows for anomaly detection using the extracted high-level semantics of the data. 


\newpage
\small
\bibliographystyle{abbrv}
\bibliography{biva_references}
\normalsize

\onecolumn
\appendix 

\section{Deep Learning and Variational Inference}\label{sec:vae_app}
\begin{wrapfigure}{R}{0.5\textwidth}
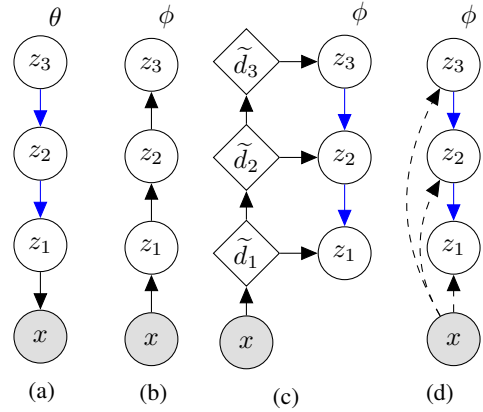

\vspace{-0.8cm}
\centering
\def\col{blue}
\begin{subfigure}{0.1\textwidth}
\centering	  
      \tikz{
     \node[obs] (x) {$x$};%
     
     \node[latent,above=.5cm of x](z1bu){$z_1$}; %
     
     \node[latent,above=.5cm of z1bu](z2bu){$z_2$}; %
     \node[latent,above=.5cm of z2bu] (z3bu) {$z_3$}; %
     \node[above=of z3bu, xshift=.2cm, yshift=-1.cm] (theta) {$\theta$}; %
     \edge[]{z1bu}{x}
     \edge[\col]{z2bu}{z1bu}
     \edge[\col]{z3bu}{z2bu}
     }
     \caption{ }\label{fig:vae_gen}
\end{subfigure}
\begin{subfigure}{0.1\textwidth}
\centering	  
      \tikz{
     \node[obs] (x) {$x$};%
     
     \node[latent,above=.5cm of x](z1bu){$z_1$}; %
     
     \node[latent,above=.5cm of z1bu](z2bu){$z_2$}; %
     \node[latent,above=.5cm of z2bu] (z3bu) {$z_3$}; %
     \node[above=of z3bu, xshift=.2cm, yshift=-1.cm] (phi) {$\phi$}; %
     \edge[]{x}{z1bu}
     \edge[]{z1bu}{z2bu}
     \edge[]{z2bu}{z3bu}
     }
     \caption{ }\label{fig:vae_inf}
\end{subfigure}
\begin{subfigure}{0.14\textwidth}
\centering
\def\col{blue}
      \tikz{
     \node[obs] (x) {$x$};%
     \node[det,above=.38cm of x] (d1){$\widetilde{d}_1$}; %
     \node[det,above=.38cm of d1] (d2){$\widetilde{d}_2$}; %
     \node[det,above=.38cm of d2] (d3){$\widetilde{d}_3$}; %

     \node[latent,right=.5cm of d1](z1td){$z_1$}; %
     \node[latent,right=.5cm of d2](z2td){$z_2$}; %
     \node[latent,right=.5cm of d3] (z3td) {$z_3$}; %
     \node[above=of z3td, xshift=.2cm, yshift=-1.cm] (phi) {$\phi$}; 
     
     %
     \edge[]{x}{d1}
     \edge[\col]{z2td}{z1td}
     \edge[]{d1}{d2}
     \edge[\col]{z3td}{z2td}
     \edge[]{d2}{d3}
     \edge[]{d1}{z1td}
     \edge[]{d2}{z2td}
     \edge[]{d3}{z3td}
     }
     \caption{ }\label{fig:lvae_inf}
\end{subfigure}
\begin{subfigure}{0.14\textwidth}
\centering
\def\col{blue}
      \tikz{
     \node[obs] (x) {$x$};%
     \node[latent,above=.5cm of x](z1td){$z_1$}; %
     \node[latent,above=.5cm of z1td](z2td){$z_2$}; %
     \node[latent,above=.5cm of z2td](z3td){$z_3$}; %
     \node[above=of z3td, xshift=.2cm, yshift=-1.cm] (phi) {$\phi$}; 
     
     %
     \edge[dashed]{x}{z1td};
     \edge[\col]{z2td}{z1td};
     \edge[\col]{z3td}{z2td};
     \edge[dashed, bend left]{x}{z2td};
     \edge[dashed, bend left]{x}{z3td};
     }
     \caption{ }\label{fig:lvae_inf_skip}
\end{subfigure}
\caption{(a) Generative model of a VAE/LVAE with $L=3$ stochastic variables, (b) VAE inference model, (c) LVAE inference model, and (d) skip connections among stochastic variables in the LVAE where dashed lines denote a skip-connection. {\color{blue} Blue} arrows indicate that there are shared parameters between the inference and generative model.}
\vspace{-0.8cm}
\label{fig:vaelvae}
\end{wrapfigure}

The introduction of stochastic backpropagation \citep{Paisley2012,Hoffman2013} and the variational auto-encoder (VAE) \citep{Kingma13,Rezende14} has made approximate Bayesian inference and probabilistic latent variable models applicable to machine learning problems considering complex data distributions, e.g. natural images, audio, and text. The VAE is a generative model parameterized by a neural network $\theta$ and is defined by an observed variable $x$ that depends on a hierarchy of stochastic latent variables $\mathbf{z}=z_1, ..., z_L$ so that: $p_{\theta}(x, \mathbf{z}) = p_{\theta}(x|z_1) p_{\theta}(z_L) \prod_{i=1}^{L-1} p_{\theta}(z_i|z_{i+1})$.
This is illustrated in Figure \ref{fig:vae_gen}.

The distributions $p_{\theta}(z_i|z_{i+1})$ over the latent variables of the VAE are normally defined as Gaussians with diagonal covariance, whose parameters depend on the previous latent variable in the hierarchy (with the top latent variable $p_{\theta}(z_L)=\mathcal{N}(z_L;0,I)$). The likelihood $p_{\theta}(x|z_1)$ is typically a Gaussian distribution for continuous data, or a Bernoulli distribution for binary data.

In order to learn the parameters $\theta$ we seek to maximize the log marginal likelihood over a training set: $\sum_i \log p_\theta(x_i) = \sum_i \log \int p_{\theta}(x_i, \mathbf{z}_i) d\mathbf{z}_i$.
However, complex data distributions require an expressive model, which makes the above integral intractable. In order to circumvent this, we use Variational Inference \citep{Jordan99} and introduce a posterior approximation $q_{\phi}(\mathbf{z}|x)$, known as \textit{inference network} or \textit{encoder}, that is parameterized by a neural network $\phi$. Using Jensen's inequality we can derive the \textit{evidence lower bound} (ELBO), a lower bound to the integral in the marginal likelihood which is a function of the variational approximation $q_{\phi}(\mathbf{z}|x)$ and the generative model $p_{\theta}(x,\mathbf{z})$:
\begin{align}
    \log p_{\theta}(x) \ge \mathbb{E}_{q_{\phi}(\mathbf{z}|x)}\left[\log \frac{p_{\theta}(x,\mathbf{z})}{q_{\phi}(\mathbf{z}|x)} \right] \equiv \mathcal{L}(\theta, \phi)\ . \label{eq:elbo_appendix}
\end{align}
The parameters $\theta$ and $\phi$ can be optimized by maximizing the ELBO with stochastic backpropagation and the reparameterization trick, which allows using gradient ascent algorithms with low variance gradient estimators \citep{Kingma13, Rezende14}. 
As illustrated in Figure \ref{fig:vae_inf}, in a VAE the variational approximation is factorized with a bottom-up structure, $q_{\phi}(\mathbf{z}|x) = q_{\phi}(z_1|x)\prod_{i=1}^{L-1} q_{\phi}(z_{i+1}|z_i)$, 
so that each latent variable is conditioned on the variable below in the hierarchy. For ease of computation, all the factors in the variational approximation are typically assumed to be Gaussians whose mean and diagonal covariance are parameterized by neural networks.

\paragraph{Latent variable collapse in VAEs.}
A deep hierarchy of latent stochastic variables will result in a more expressive model. However, the additional variables come at a price. As shown in \citep{Chen2017,Maaloe17}, we can rewrite the ELBO (eq. (\ref{eq:elbo})):
\begin{align}
    \mathcal{L}(\theta,\phi) =  \mathbb{E}_{q_{\phi}(\mathbf{z}|x)}\left[ \log \frac{p_{\theta}(x,z_{<L}|z_L)}{q_{\phi}(z_{<L}|x)}\right] - \nonumber \mathbb{E}_{q_{\phi}(z_{<L}|x)}\left[KL[q_{\phi}(z_L|z_{<L})||p_{\theta}(z_L))]\right] \ .
\end{align}
From the above, it becomes obvious that, during the optimization of the VAE, the top stochastic latent variables may have a tendency to \textit{collapse} into the prior, i.e. $q_{\phi}(z_L|z_{<L})=p_{\theta}(z_L)=\mathcal{N}(z_L; 0,I)$, if the model $p_{\theta}(x,z_{<L}|z_L)$ is powerful enough. This is supported by empirical results in \citep{Sonderby2016, Bowman2015} amongst others. The tendency has limited the applicability of deep VAEs in problems with complex data distributions, and has pushed VAE research towards the extension of shallow VAEs with autoregressive models, that allow capturing a \textit{lossy} representation in the latent space while achieving strong generative performances \citep{Gulrajani2016, Chen2017}. Another research direction has focused on learning more complex prior distributions through normalizing flows \citep{Rezende2015,Tomczak16, Kingma2016}. Our research considers instead the original goal of building expressive models that can exploit a deeper hierarchy of stochastic latent variables while avoiding variable collapse.

\section{Detailed Model Description}\label{app:model_description}
\begin{figure*}[t]
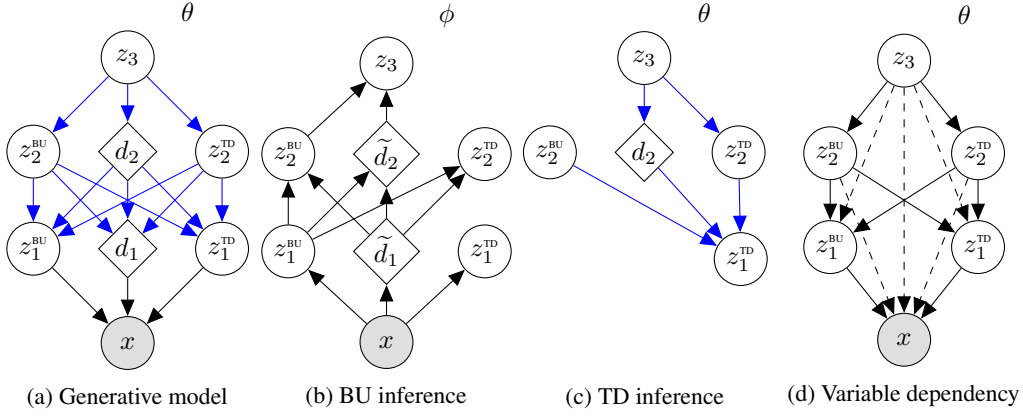

\begin{subfigure}{0.24\textwidth}
\centering	  
\def\col{blue}
      \tikz{
     \node[obs] (x) {$x$};%

     \node[det,above=.5cm of x] (d1){$d_1$}; %
     \node[latent,right=0.5cm of d1] (z1td){$z_1^{\scalebox{0.4}{TD}}$}; %
     \node[latent,left=0.5cm of d1](z1bu){$z_1^{\scalebox{0.4}{BU}}$}; %
     
     \node[det,above=.5cm of d1] (d2){$d_2$}; %
     \node[latent,right=0.5cm of d2] (z2td){$z_2^{\scalebox{0.4}{TD}}$}; %
     \node[latent,left=0.5cm of d2](z2bu){$z_2^{\scalebox{0.4}{BU}}$}; %
     \node[latent,above=.5cm of d2] (z3td){$z_3$}; %
     
     \node[above=of z3td, xshift=.8cm, yshift=-1.cm] (theta) {$\theta$}; %
     
     \edge[\col]{z3td}{d2}
     \edge[\col]{z3td}{z2bu}
     \edge[\col]{z3td}{z2td}
     
     \edge[\col]{d2}{d1}
     \edge[\col]{d2}{z1bu}
     \edge[\col]{d2}{z1td}
     
     \edge[\col]{z2td}{d1}
     \edge[\col]{z2td}{z1bu}
     \edge[\col]{z2td}{z1td}

     \edge[\col]{z2bu}{d1}
     \edge[\col]{z2bu}{z1bu}
     \edge[\col]{z2bu}{z1td}
     
     \edge[]{d1}{x}
     \edge[]{z1td}{x}
     \edge[]{z1bu}{x}

     }
     \caption{Generative model}\label{fig:lvae++_gen}
\end{subfigure}
\begin{subfigure}{0.24\textwidth}
\centering	  
\def\col{blue}
      \tikz{
     \node[obs] (x) {$x$};%

     \node[det,above=.4cm of x] (d1){$\widetilde{d}_1$}; %
     \node[latent,right=0.5cm of d1] (z1td){$z_1^{\scalebox{0.4}{TD}}$}; %
     \node[latent,left=0.5cm of d1](z1bu){$z_1^{\scalebox{0.4}{BU}}$}; %
     
     \node[det,above=.4cm of d1] (d2){$\widetilde{d}_2$}; %
     \node[latent,right=0.5cm of d2] (z2td){$z_2^{\scalebox{0.4}{TD}}$}; %
     \node[latent,left=0.5cm of d2](z2bu){$z_2^{\scalebox{0.4}{BU}}$}; %
     \node[latent,above=.4cm of d2] (z3td){$z_3$}; %
     
     \node[above=of z3td, xshift=.8cm, yshift=-1.cm] (theta) {$\phi$}; %
     
     \edge[]{d2}{z3td}
     \edge[]{z2bu}{z3td}

     \edge[]{d1}{d2}
     \edge[]{z1bu}{d2}

     \edge[]{d1}{z2td}
     \edge[]{z1bu}{z2td}

     \edge[]{d1}{z2bu}
     \edge[]{z1bu}{z2bu}

     \edge[]{x}{d1}
     \edge[]{x}{z1td}
     \edge[]{x}{z1bu}

     }
     \caption{BU inference}\label{fig:lvae++_inf_bu}
\end{subfigure}
\begin{subfigure}{0.24\textwidth}
\centering	  
\def\col{blue}
      \tikz{
     \node[] (x) {};%

     \node[above=1.1cm of x] (d1){}; %
     \node[latent,right=0.8cm of d1] (z1td){$z_1^{\scalebox{0.4}{TD}}$}; %
     \node[left=0.5cm of d1](z1bu){}; %
     
     \node[det,above=.9cm of d1] (d2){$d_2$}; %
     \node[latent,right=0.5cm of d2] (z2td){$z_2^{\scalebox{0.4}{TD}}$}; %
     \node[latent,left=0.5cm of d2](z2bu){$z_2^{\scalebox{0.4}{BU}}$}; %
     \node[latent,above=.5cm of d2] (z3td){$z_3$}; %
     
     \node[above=of z3td, xshift=.8cm, yshift=-1.cm] (theta) {$\theta$}; %
     
     \edge[\col]{z3td}{d2}
     \edge[\col]{z3td}{z2td}
     
     \edge[\col]{d2}{z1td}
     
     \edge[\col]{z2td}{z1td}

     \edge[\col]{z2bu}{z1td}

     }
     \caption{TD inference}\label{fig:lvae++_inf_td}
\end{subfigure}
\begin{subfigure}{0.24\textwidth}
\centering	  
\def\col{blue}
      \tikz{
     \node[obs] (x) {$x$};%

     \node[above=.75cm of x] (d1){}; %
     \node[latent,right=0.5cm of d1] (z1td){$z_1^{\scalebox{0.4}{TD}}$}; %
     \node[latent,left=0.5cm of d1](z1bu){$z_1^{\scalebox{0.4}{BU}}$}; %
     
     \node[above=1.cm of d1] (d2){}; %
     \node[latent,right=0.5cm of d2] (z2td){$z_2^{\scalebox{0.4}{TD}}$}; %
     \node[latent,left=0.5cm of d2](z2bu){$z_2^{\scalebox{0.4}{BU}}$}; %
     \node[latent,above=.75cm of d2] (z3td){$z_3$}; %
     
     \node[above=of z3td, xshift=.8cm, yshift=-1.cm] (theta) {$\theta$}; %
     
     \edge[]{z3td}{z2bu}
     \edge[]{z3td}{z2td}
     \edge[dashed]{z3td}{z1bu}
     \edge[dashed]{z3td}{z1td}
     \edge[dashed]{z3td}{x}

     \edge[]{z2td}{z1bu}
     \edge[]{z2td}{z1td}
     \edge[dashed]{z2td}{x}

     \edge[]{z2bu}{z1bu}
     \edge[]{z2bu}{z1td}
     \edge[dashed]{z2bu}{x}

     \edge[]{z1td}{x}
     \edge[]{z1bu}{x}

     }
     \caption{Variable dependency}\label{fig:lvae++_gen_skip}
\end{subfigure}
\caption{A $L=3$ layered \nm with (a) the generative model, (b) bottom-up (BU) inference path, (c) top-down (TD) inference path, and (d) variable dependency of the generative models where dashed lines denote a skip-connection.  {\color{blue} Blue} arrows indicate that the deterministic parameters are shared within the generative model or between the generative and inference model.}
\label{fig:lvae++detailed}
  \vspace*{-0.4cm}
\end{figure*}

\paragraph{Generative model.} The generative model (see Figure \ref{fig:lvae++_gen}) has a top-down path going from $z_L$ through the intermediary stochastic latent variables to $x$. Between each stochastic layer there is a ResNet block with $M$ layers set up similarly to \citep{Salimans17}. Weight normalization \citep{Salimans2016a} is applied in all neural network layers. In the generative model, the BU and TD units are not distinguished so we write $z_i =(z_i^{\scalebox{0.6}{BU}},z_i^{\scalebox{0.6}{TD}})$. We use $f_{i,j}$ to denote the neural network function (a function of generative model parameters $\theta$) of ResNet layer $j$ associated with stochastic layer $i$. The feature maps are written as $d_{i,j}$. The generative process can then be iterated as $z_L \sim \mathcal{N}(0,I)$ and $i=L-1,L-2,\ldots,1$:
\begin{align}
    d_{i,0} &= z_{i+1} \\
    d_{i,j} &= <f_{\theta_{i,j}}(d_{i,j-1}); d_{i+1,j}> \text{ \textbf{for} } j=1,...,M \label{eq:gen_path}\\
    z_{i} &= \mu_{\theta,i}(d_{i,M}) + \sigma_{\theta,i}(d_{i,M}) \otimes \epsilon_i\ ,
\end{align}
where $d_{L,j}=0$, $<;>$ denotes concatenation of feature maps in the convolutional network and hidden units in the fully connected network, $\epsilon \sim \mathcal{N}(0,I)$ and $\mu(\cdot)$ and $\sigma(\cdot)$ are parameterized by neural networks. 
To complete the generative model $p(x|\mathbf{z})$ is written in terms of $z_1$ and $d_1$ through a ResNet block $f_{0}$. 

\paragraph{Inference model.} The inference model (see Figure \ref{fig:lvae++_inf_bu} and \ref{fig:lvae++_inf_td}) consists of a bottom-up (BU) and top-down (TD) paths such that bottom-up stochastic units only receive bottom-up information whereas the top-down units receive both bottom-up and top-down information. The top-down path shares parameters with the generative model. For each stochastic latent variable $z_i$ in $i=1,...,L$ we use a ResNet block with $M$ layers and there are associated neural network functions $g_{i,j}$, $j=1,\ldots,M$ with parameters collectively denoted by $\phi$. 
The deterministic feature map of layer $i,j$ is denoted by $\tilde{d}_{i,j}$:
\begin{align}
    \tilde{d}_{i,0} &= \left\{ \begin{matrix} x & i=1\\ < z_{i-1}; \tilde{d}_{i-1,M} > & \mathrm{otherwise} \end{matrix} \right. \\
    \tilde{d}_{i,j}& = <g_{i,j}(\tilde{d}_{i,j-1}); \tilde{d}_{i-1,j}> \text{ \textbf{for} } j=1,...,M\ ,\label{eq:td_path}\\
     z_{i}^{\scalebox{0.6}{BU}} &= \mu^{\scalebox{0.6}{BU}}_{i}(\tilde{d}_{i,M}) + \sigma^{\scalebox{0.6}{BU}}_{i}(\tilde{d}_{i,M}) \otimes \epsilon^{\scalebox{0.4}{BU}}_i \label{eq:bu_path} 
\end{align}
where $\epsilon \sim \mathcal{N}(0,I)$. Finally, to infer the top-down latent we use the bottom-up latent $z_{i}^{\scalebox{0.6}{TD}}$ inferred in eq.\ (\ref{eq:bu_path}) and pass them through the generative path eq.\ (\ref{eq:gen_path}) for $i=L-1,L-2,\ldots,2$ to determine $d_{i,M}$ and 
\begin{align}
      z_{i}^{\scalebox{0.6}{TD}} &= \mu^{\scalebox{0.6}{TD}}_{i}(< \tilde{d}_{i,M}; d_{i,M} >) + \sigma^{\scalebox{0.6}{TD}}_{i}(< \tilde{d}_{i,M}; d_{i,M} >) \otimes \epsilon^{\scalebox{0.4}{TD}}_i \ . \label{eq:td_path}
\end{align}

\section{Experimental Setup}\label{app:experimental_setup}
Throughout all experiments, we follow the \nm model description that is described in detail in Appendix \ref{app:model_description} and \ref{app:semi_supervised}.

\paragraph{Optimization.} All models are optimized using Adamax \citep{Kingma14a} with a hyperparameter setting similar to the one used in \citep{Kingma2016}. They are trained with a batch-size of 48 where the binary image experiments are trained on a single GPU and the natural image experiments are trained on two GPUs (by splitting the batch in 2 and then taking the mean over the gradients). For evaluation, we use exponential moving averages of the parameters space, similar to \citep{Kingma2016,Salimans17}.

\paragraph{Binary image architecture.} \nm has $L=6$ layers. The $g_{\phi_1}$ neural networks are defined by $M=3$, 64x5x5 (number of kernels x kernel width x kernel height) convolutional layers and an overall stride of 2. Neural networks $i=2,...,6$ are defined by four $M=3$, 64x3x3 convolutional layers. The final neural network, $i=6$, applies a stride of 2. All stochastic latent variables are densely connected layers of dimension $48, 40, 32, 24, 16, 8$ for $1, ..., L$ respectively. We apply a dropout rate of $0.5$ for both the deterministic layers in the generative as well as the inference model.

\paragraph{Natural image architecture (32x32).} \nm has $L=15$ layers. The $g_{\phi_1}$ neural networks are defined by $M=3$, 96x5x5 convolutional layers and an overall stride of 2. Neural networks $i=2,...,15$ are defined by $M=3$, 96x3x3 convolutional layers. Neural networks $11$ and $15$ are defined with a stride of 2. All stochastic latent variables are parameterized by convolutional layers with $38, 36, 34, ..., 10$ feature maps for $1, 2, 3, ..., L$ respectively. The kernel width and height of the stochastic latent variables are defined similarly to the dimension of the subsequent output after striding. We apply a dropout rate of $0.2$ in the deterministic layers of the inference model.

\paragraph{Natural image architecture (64x64).} \nm has $L=20$ layers. The $g_{\phi_1}$ and $g_{\phi_2}$ neural networks are defined by $M=3$, 64x7x7 and 64x5x5 convolutional layers respectively with a stride of 2 in each. Neural networks $i=3,...,11$ are defined by $M=3$ 64x3x3 convolutional layers. Neural network $11$ is defined with a stride of 2. Neural networks $i=12,...,20$ are defined by $M=3$, 128x3x3 convolutional layers and network $20$ has a stride of 2. All stochastic latent variables are parameterized by convolutional layers with $20, 19, 18, ..., 1$ feature maps for $1, 2, 3, ..., L$ respectively. The kernel width and height of the stochastic latent variables are defined similarly to the dimension of the subsequent output after striding. We apply a dropout rate of $0.2$ in the deterministic layers of the inference model.

\section{Modeling Complex 2D Densities}\label{app:2d_density}

\begin{table}[h!]
\renewcommand\figurename{Table}
\begin{center}
\begin{small}
\begin{sc}
\begin{tabular}{l c}
& Potential $U(\mathbf{Z})$ \\
\hline
\textbf{1:} & $ \frac{1}{2}\left(\frac{\|\mathbf{z}\|-2}{0.4}\right)^{2}- \ln \left( e^{-\frac{1}{2}\left[\frac{\mathbf{Z}_{1}-2}{0.6}\right]^{2} } +e^{-\frac{1}{2}\left[\frac{\mathbf{Z}_{1}+2}{0.6}\right]^{2}} ) \right)$   \\
\textbf{2:} &  $\frac{1}{2}\left[\frac{\mathbf{Z}_{2}-w_{1}(\mathbf{Z})}{0.4}\right]^{2}$ \\
\textbf{3:} &  $-\ln \left(e^{-\frac{1}{2}}\left[\frac{\mathbf{Z}_{2}-\boldsymbol{w}_{1}(\mathbf{Z})}{0.35}\right]^{2}+e^{-\frac{1}{2}\left[\frac{\mathbf{Z}_{2}-\boldsymbol{w}_{1}(\mathbf{Z})+w_{2}(\mathbf{Z})}{0.35}\right]^{2} )}\right) $ \\
\textbf{4:} &  $ -\ln \left(e^{-\frac{1}{2}\left[\frac{\mathbf{Z}_{2}-w_{1}(\mathbf{Z})}{0.4}\right]^{2}}+e^{-\frac{1}{2}\left[\frac{\mathbf{Z}_{2}-w_{1}(\mathbf{Z})+w_{3}(\mathbf{Z})}{0.35}\right]^{2} )}\right) $ \\
\hline
& with $w_{1}(\mathbf{z})=\sin \left(\frac{2 \pi \mathbf{z}_{1}}{4}\right)$, $w_{2}(\mathbf{z})=3 e^{-\frac{1}{2}\left[\frac{\left(\mathbf{Z}_{1}-1\right)}{0.6}\right]^{2}}$, \\
& $w_{3}(\mathbf{z})=3 \sigma\left(\frac{\mathbf{Z}_{1}-1}{0.3}\right)$ and $\sigma(x)=1 /\left(1+e^{-x}\right)$ \ . \\
\hline
\end{tabular}
\end{sc}
\end{small}
\end{center}

\caption{Potentials defining the target densities $p(\mathbf{z}) = \frac{e^{-U(\mathbf{z})}}{Z}$.}\label{tab:target_densities}
\end{table}

\paragraph{Problem.} 

\cite{Maaloe2016} showed that Variational Auto-Encoders can fit complex posterior
distributions for the latent space using the inference model $q_\phi(z|x)$, parameterized as a fully factorized Gaussian and $p(x)$ being a simple diagonal Gaussian. In table \ref{tab:target_densities}, we define complex non-Gaussian densities using a potential model $U(\mathbf{Z})$, as described in \cite{Rezende2015}. While modeling such distributions remains within the reach of an adequately complex Variational Autoencoder, optimizing such a model remains challenging.

\paragraph{Objective.} 

Similarly to \cite{Maaloe2016}, we choose $p(x)$ to be an isotropic Gaussian and we model the target density using the top stochastic variable: $p(z_L) = \frac{e^{-U(z)}}{Z}$. This results in the following bound:

\begin{align}
\log p(x) \geq \mathbb{E}_{q_{\phi}(x, \mathbf{z})} \left[ \log  \frac{p_{\theta}(x|z_1)p(z_L)}{q_{\phi}(x)} + \sum_{i=1}^{L-1} \log \frac{p_{\theta}(z_i | z_{i+1})}{q_{\phi}(z_{i,TD}|z_{i+1},x)q_{\phi}(z_{i+1}|z_{i,BU}, x)}\right] \ .
\end{align}

\paragraph{Experimental Setup.}

We test \nm against the VAE and LVAE models using the same number of stochastic variables, hence the models use the same number of intermediate layers. All models are implemented using 5 stochastic layers, MLPs with one hidden layer of size $128$ and with residual connections. The chosen architecture is voluntary kept minimal, therefore the task remains challenging for all models.

We train all models for $1e^{4}$ iterations using the Adamax optimizer. We use batch sizes of size $512$. The potential is linearly annealed from $0.1$ to $1$ during $5e^{3}$ steps. In order to avoid posterior collapse, $0.5$ \textit{freebits} are applied to each stochastic layer. The learning rate is linearly increased from $1e^{-5}$ to $3e^{-3}$ and exponentially annealed back to $1e^{-5}$.

In order to measure the quality of the posterior density, we estimate $KL(q(z_L) || p(z_L))$ using $1e^{6}$ posterior samples evaluated using a grid of size $(-2,2)^2$ with a resolution of $100 \times 100$. Each model is trained $100$ times for each density.

\paragraph{Results.}

\begin{figure}[!h]
\centering
\includegraphics[width=1.0\textwidth]{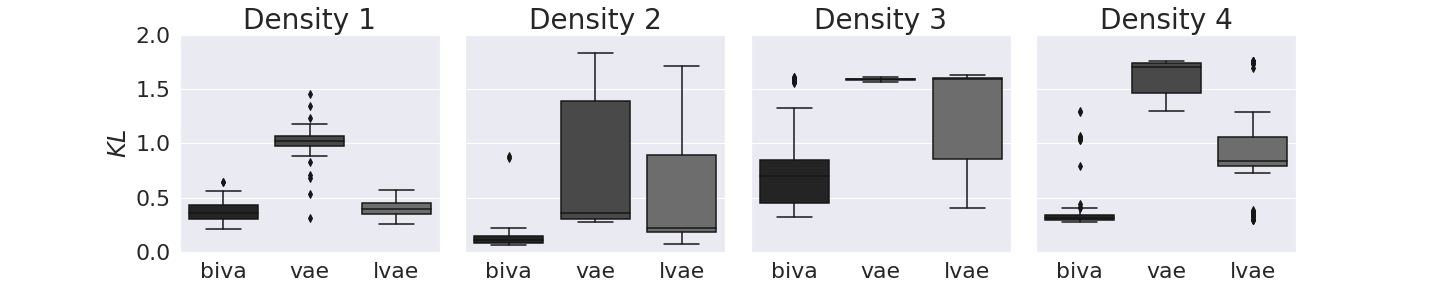}
\caption{Distribution of the $KL(q(z_L) || p(z_L)))$ estimate for each model, each target density $p(z_L)$ and for different initial random seeds. We collected 100 runs for each model and for each density. We found that \nm behaves more consistently and often yield better approximations than the baseline models.}\label{fig:2ddensity_bar}
\end{figure}

According to the approximate $KL(q(z_L) || p(z_L))$, we found that \nm tends to learn a posterior that lies closer to the target density. Figure \ref{fig:2ddensity_bar} shows that \nm often learns more complex features than the baseline models, which posteriors remain closer to the modes. Figure \ref{fig:2ddensity_bar} reveals that LVAE is able to find solutions that are competitive with the best \nm samples according to $KL(q(z_L) || p(z_L))$. However, this happens very rarely whereas \nm has a more robust optimization behaviour. 

\begin{figure}[!h]
\centering
\includegraphics[width=.8\textwidth]{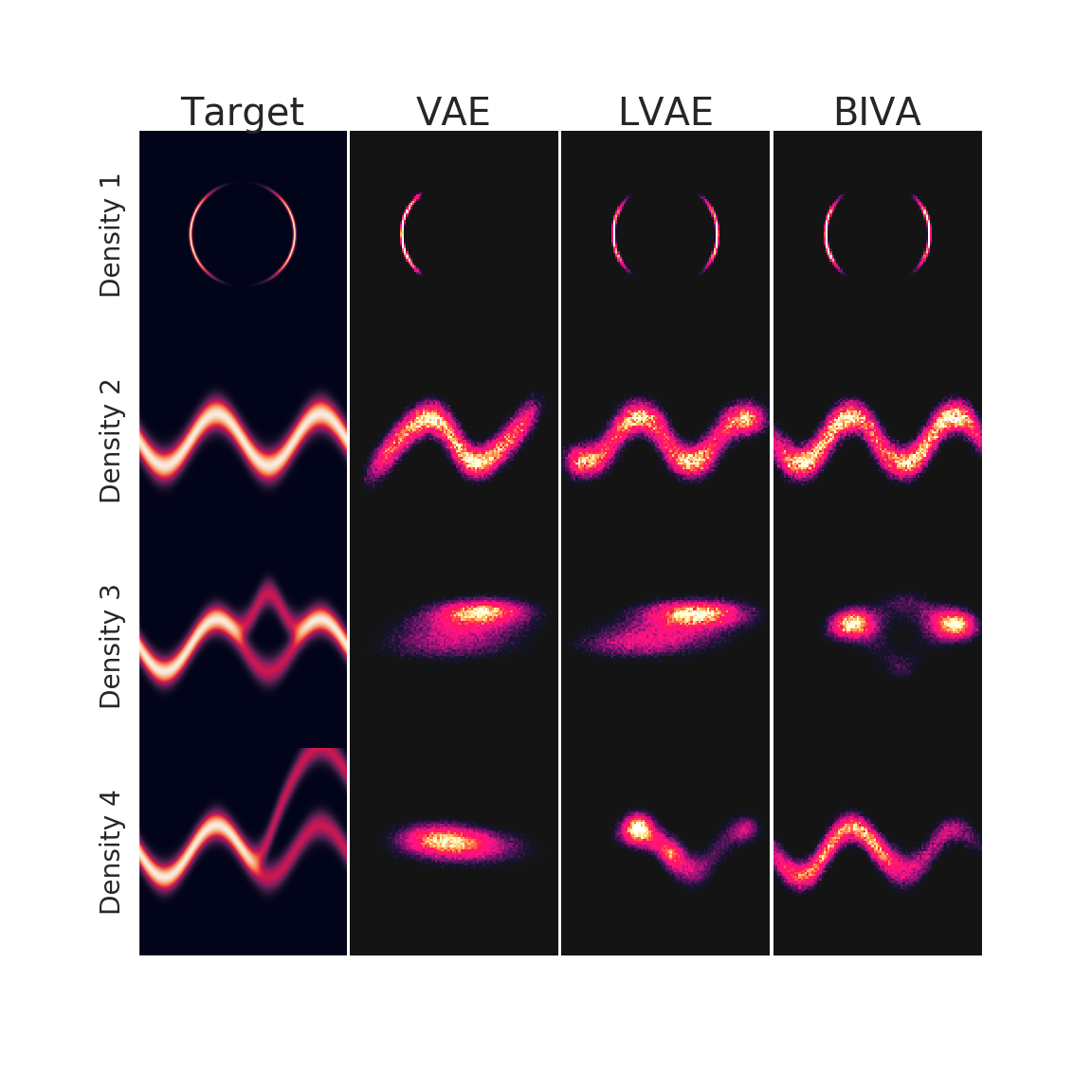}
\caption{Target densities $p(z_L)$ and the median posterior distributions $q(z_L)$ for each model according to $KL(q(z_L) || p(z_L)))$ out of 100 runs for each model and for each density.}\label{fig:2ddensity_agg}
\end{figure}

\section{Initial Results on Text Generation Tasks}\label{app:text_modeling}

\begin{table}[h!]
\renewcommand\figurename{Table}
\begin{center}
\begin{small}
\begin{sc}
    \begin{tabular}{l c c c c}
     & parameters & $-\log p(x)$ & kl & ppl\\
    \hline
    \textit{Results with autoregressive components, no dropout} \\
    LSTM & $15.0M$ & $=41.49$ & $-$ & $36.28$ \\
    RNN-VAE \cite{Bowman2015}, $\mathcal{L}_1$, warmup & $23.7M$ & $\le 42.09$ & $1.61$ & $38.21$ \\
    RNN-VAE \cite{Bowman2015}, $\mathcal{L}_1$, finetuned & $23.7M$ & $\le 42.41$ & $5.13$ & $39.26$\\
    Hybrid VAE \cite{semeniuta2017hybrid}, $\mathcal{L}_1$, finetuned & $23.7M$ &$\le 42.24$ & $4.67$ & $38.70$ \\
    \textbf{\nm} L=7, $\mathcal{L}_1$, finetuned & $23.0M$ & $\le 42.34$ & $10.15$ & $39.04$\\
    \hline
     \textit{Results without autoregressive components, no dropout} \\
    Hybrid VAE \cite{semeniuta2017hybrid}, $\mathcal{L}_1$, finetuned & $15.0M$ & $\le 54.53$ & $ 14.10$ & $112.1$\\
    \textbf{\nm} L=7 Finetuned, $\mathcal{L}_1$ & $14.0M$ & $\le 54.13$ & $15.33$ & $108.3$\\
    \hline
    \end{tabular}%
\end{sc}
\end{small}
\end{center}

\caption{Test performances on the BookCorpus with 1 importance weighted sample (sentences limited to 40 words). The RNN-VAE and Hybrid VAE are are trained and evaluated from our own implementation.}\label{table:bookcorpus-results}
\end{table}

Optimizing generative models coupled with autoregressive models is a difficult task. Such coupling causes the posterior to collapse, and the latent variables are ignored. Nonetheless, autoregressive components remain a cornerstone of the generative models for text \cite{Bowman2015, semeniuta2017hybrid,shah2018generating}. In order to enforce the model to use the latent variable, previous efforts aimed at weakening the decoder using powerful regularizing \textit{tricks}, such as word dropout \citep{Bowman2015}. We investigate the use of \nm in the context of sentence modeling without weakening the decoder. We show that it allows optimizing the latent variables more effectively, resulting in a higher measured KL when compared to the RNN-VAE \citep{Bowman2015} and the Hybrid VAE \citep{semeniuta2017hybrid}.

\paragraph{Dataset.} We use the Bookcorpus dataset \cite{bookcorpus} of sentences of maximum 40 words, no preprocessing is performed and sentences are tokenized using the white spaces. We defined a vocabulary of $20000$ words and filtered out the sentences that contain non-indexed tokens. We randomly sampled $10000$ sentences for testing and used the remaining $56$M sentences for training.  

\paragraph{Models.} We couple \nm with an LSTM decoder, using the output of the convolutional model as an input sequence for the auto-regressive model. We compare our model against a LSTM language model \cite{hochreiter1997long}, the RNN-VAE \cite{Bowman2015}, and the Hybrid VAE \cite{semeniuta2017hybrid}, which couples a convolutional architecture with an LSTM decoder. We also perform experiments without using autoregressive components.

All LSTM models are parameterized by $1024$ units and we use embeddings of dimension $512$. This results in an RNN-VAE model with 23.7M parameters and we limit the other models to use the same total number of parameters. This results in using a limited number of stochastic layers for the \nm and small a small number of kernels of $128$.

\paragraph{Training.} We trained the models for $5$ epochs with an initial learning rate of $2e^{-3}$ using the Adamax optimizer. We used batches of size $512$ and used only one stochastic sample. We train all latent variable models using the \textit{freebits} method from \cite{Kingma2016} with an initial KL budget of $30$ nats distributed equally over the stochastic variables and we incrementally decrease the \textit{freebits} value \textit{on plateau}. We also train the RNN-VAE baseline using the deterministic warmup method \citep{Bowman2015,Sonderby2016} for comparison.

\paragraph{Likelihood and latent variables usage.}
We report the test set results in table \ref{table:bookcorpus-results} and test samples in \ref{table:text-samples} and reconstructions in table \ref{table:text-reconstructions}. While \nm without the autoregressive decoder is not competitive with an LSTM language model, we observe that replacing the LSTM inference model by a \nm model allows exploiting the latent space more actively, which results in a higher measured KL than the RNN-VAE and Hybrid VAE baselines.

\begin{table}[!h]
\renewcommand\figurename{Table}
\begin{center}
\resizebox{\textwidth}{!}{%
\begin{small}

\begin{tabular}{ll}
\toprule
                                                                                                         BIVA+LSTM &                                                                                                                    RNN-VAE \\
\midrule
                                                                    he said .                                      &                                                                              `` two .                                      \\
  i tried to think of something to say to him , but he was already on his way back to the house .                  &                                                           `` you do n't have to do this . ''                               \\
                                     it sounded as if he was going to say something .                              &                       the light from the lamp was dim , but the light was dim and the room was dark .                      \\
                                              `` and that 's why you 're coming . ''                               &                                                         or a nuclear bomb , or something .                                 \\
                                                                  `` what ? ''                                     &                                                                        `` the baby ? ''                                    \\
                                                              she swallowed .                                      &                                                         `` you 're not going to kill me . ''                               \\
                                                              `` i want you . ''                                   &                                                                   she was n't going to .                                   \\
                glancing up , i saw the way he was staring at me with a look of pure hatred .                      &                                        `` i guess we could have been more careful , '' he said .                           \\
                                                             i need a favor . ''                                   &                                                   there are some things that are not good .                                \\
                                                                  he did n't .                                     &                                                                   `` you 're a good man .                                  \\
                                                             you 're not dead .                                    &                                                          i had n't been able to get it out .                               \\
                                                      i stood , and he followed .                                  &                                            `` you 're going to have to be careful , '' he said .                           \\
                                              `` can i sit on the couch and talk ? ''                              &                                                                    it 's not a bad idea .                                  \\
                     `` it was n't until i was fifteen , i was n't in the mood to be around .                      &                                                                            he asked .                                      \\
                                                        i looked down at my lap .                                  &                                    `` this is a bad idea , '' he said , his voice a little hoarse .                        \\
                                                       the smile disappeared .                                     &                                                         `` i 'm sure he 's in love with you .                              \\
                                     it was hard to tell which one was more of a rock .                            &   as he stepped out of the car , he saw the man standing in the doorway , his eyes wide and his face pale .                \\
                                                   i 'm not sure it 's a good idea .                               &                                                                                   .                                        \\
                                                               the first two .                                     &                                                                               `` no .                                      \\
                                                                he was there .                                     &                                        `` in the meantime , i need to get some sleep , '' i said .                         \\
                                                      `` all of you , '' joe said .                                &                                                                            i was n't .                                     \\
                                            he did n't care if he was n't a vampire .                              &                                                                did i want to talk to you ?                                 \\
                                            her mouth curved up , then she nodded .                                &                                                            `` i want to hear you say it . ''                               \\
                                             just tell me what you want in the end .                               &                                                    the train was already in the driveway .                                 \\
                                                                  and again .                                      &                                                                             `` good .                                      \\
                                      the other man 's voice was hoarse and ragged .                               &                                 i just needed to get out of here , and i needed to get out of here .                       \\
          i had n't known that was a bad idea , but i had n't been able to get it out of my head .                 &                                                                  `` this is a good idea .                                  \\
                                    your brother is the most important thing to me .                               &                                                                            `` hey . ''                                     \\
                                          you dont need to go to the police , right ?                              &                                                      she took a deep breath and let it out .                               \\
                                                      there was a long silence .                                   &                                                                    then he kissed her .                                    \\
                                                                 i looked up .                                     &                          i felt a warm hand on my shoulder and a warm smile spread across my face .                        \\
                he nodded , and he looked at me , and i could tell he was thinking about it .                      &                                                                       `` he 's dead . ''                                   \\
                                                             `` hang on , baby .                                   &                                        at the time , i was going to have to get out of the house .                         \\
                       we had to be close to the city , and we could n't afford to be here .                       &                                                      he was so close to the edge of the bed .                              \\
                             you know , it would be better if you were n't so stupid . ''                          &                                                                       `` i do n't know .                                   \\
                                                                  excuse me ?                                      &                                                              `` i do n't have a choice . ''                                \\
                                                you know how much i love you , too .                               &                                             i know i 'm not going to let him touch me , but i do .                         \\
                                       a woman 's voice , a voice that was familiar .                              &                               i could n't see the face of the man who 'd just been in the doorway .                        \\
    i have a very important business to attend to , and i 'm going to have to make a decision .                    &                          in the end , we all know that we are not going to be able to get out of this .                    \\
                          they sat on the small wooden table in the center of the room .                           &                                                                              `` yes .                                      \\
                                                              `` it 's fine . ''                                   &                                                            `` what are you doing here ? ''                                 \\
                                                      she felt a rush of relief .                                  &                                         so the only thing that mattered was that he was here .                             \\
                                                              maria , he says .                                    &                                                                 neither of them spoke .                                    \\
                                                                      what ?                                       &                                              from now on , you will be able to get out of here .                           \\
                                      `` it does n't seem like a lot to me , '' he said .                          &                                    the thought of having to kill him made him want to kill her .                           \\
                                                     he 'd told her everything .                                   &                                           the other two were staring at me , their eyes wide .                             \\
                                                            `` she 's in shock .                                   &                                  i did n't want to be a part of it , but i was n't going to let it go .                    \\
              `` after all , '' he murmured , `` i 'm going to go get the rest of the stuff . ''                   &                                                          `` i do n't want to talk about it .                               \\
                                               and then , finally , she 'd done it .                               &                                                         she looked at him , her eyes wide .                                \\
                                    her words were a whisper , but it was n't enough .                             &                                                        `` that 's what you 're going to do .                               \\
\bottomrule
\end{tabular}

\end{small}
}
\end{center}
\caption{Samples decoded from the prior of the \nm with LSTM decoder and baseline RNN-VAE.} \label{table:text-samples}
\end{table}

\begin{landscape}
\begin{table}[!h]
\renewcommand\figurename{Table}
\begin{center}
\resizebox{\textwidth}{!}{%
\begin{small}

\begin{tabular}{lll}
\toprule
                                                                                               input &                                                                                          BIVA+LSTM &                                                                                        RNN-VAE \\
\midrule
                    `` a sad song , being sung alone in the basement . ''                            &                      `` it sounds like you 've been through a lot . ''                             &                        `` you 're going to be a great father . ''                              \\
               more often , though , wherever she sank , beck was there .                            &                      more than anything , she wanted to be with him .                              &                       in the end , we all knew what was going on .                             \\
                    he looked just about as pale as i had ever seen him .                            &                               he 's still a lot more than a friend .                               &                                he was n't going to let her go .                                \\
              caleb turned and shoved him back as he took his true form .                            &              he lifted me up , his arms still wrapped around my waist .                            &   he was standing in the doorway , his hands folded in front of him .                          \\
              i gasped , tried to pull away , squeezed my legs together .                            &                       i gasped , and he was n't able to stop himself .                             &              i felt my body tense , and i could n't help but smile .                           \\
              i agreed as i adjusted myself and sat heavily in my chair .                            &                  i tried to ignore it , but my eyes were still closed .                            &                         i did n't want to be the one to tell him .                             \\
                      you bind me , UNK in darkness , though , in light .                            &                 he 'd decided to take her home , to make her feel safe .                           &                         he was more than willing to let her go .                               \\
          they promise me things , ask me questions , whisper and plead .                            &           they might be able to do something about it , but they do n't .                          &                    `` we need to talk , '' he said , his voice low .                           \\
                    i glowed as i held the bear , almost bigger than me .                            &                    i started to close my eyes , but he was too strong .                            &                   i could n't help but smile at the sight of her .                             \\
                    i wonder how much he pays them to be his guard dogs .                            &                           i had to admit that it was n't a good idea .                             &                              i do n't want to be a part of this .                              \\
                 `` hmmm , '' richard muttered , and headed up the path .                            &                     `` jesus , '' he said , his voice barely audible .                             &                            `` but you 're going to be a father .                               \\
                      he was happy that he had found it in the UNK hall .                            &                                he was n't going to be the one to go .                              &                       he was n't sure if he was going to make it .                             \\
                         at the shack , at the condo , at the hangar . ''                            &   at the moment , the only thing that mattered was that he was n't alone .                         &                     he was staring at the floor , his eyes wide .                              \\
                             `` i 'd pop to go to the dance with you . ''                            &                               `` i 'd prefer to go to the hospital .                               &                            `` i 'm going to go to the bathroom .                               \\
                  someday , i 'll share them with the rest of the world .                            &                                     and now i have a lot of my own .                               &                 `` we 're going to have to do something about it .                             \\
                     `` maybe i 'm not the right person for this one '' .                            &                        `` maybe we can get a little more of a ride . ''                            &             `` i do n't think you 're going to be able to do that . ''                         \\
                        `` gin is my sister , and she 's coming with me .                            &                  `` there 's a chance i can get a little more sleep . ''                           &                            `` if you want to , i 'll be there . ''                             \\
  thick desire stormed her ... along with a bittersweet curl of emotion .                            &                   the tension was gone , and he was n't looking at me .                            &                     the air smelled of stale cigarette smoke .                                 \\
            they caused him to stagger back and drove him to the ground .                            &                               they had to be at the top of the hill .                              &                          he 'd found a way to get her to safety .                              \\
                          you 're not much of a friar , friar , he says .                            &                   you 're not supposed to be around here , are you ? ''                            &                you 're not going to be able to do that , are you ? ''                          \\
\bottomrule
\end{tabular}

\end{small}
}
\end{center}
\caption{Reconstruction of samples from the test set using the \nm with LSTM decoder and the RNN-VAE baseline. The samples are decoded from the posterior distribution by using greedy decoding, without teacher forcing.}
\label{table:text-reconstructions}
\end{table}
\end{landscape}

%
%

\section{Semi-Supervised Learning}\label{app:semi_supervised}
When defining \nm for semi-supervised classification tasks we follow the approach described for the M2 model in \citep{Kingma14}. In addition to \nm, described in detail in Appendix \ref{app:model_description}, we introduce a classification model $q_{\phi}(y|x,z_{<L}^{\scalebox{.6}{BU}})$ in the inference model, where $y$ is the class variable, and a Categorical latent variable dependency in the generative model.

\paragraph{Inference model.} For the classification model we introduce another deterministic hierarchy with an equivalent parameterization as $\tilde{d}_{i,1}, ..., \tilde{d}_{i,M}$. We denote the hierarchy $\tilde{d}^{\scalebox{0.4}{C}}_{i,1}, ..., \tilde{d}^{\scalebox{0.4}{C}}_{i,M}$. The forward-pass is performed by:
\begin{align}
    \tilde{d}^{\scalebox{0.6}{C}}_{i,0} &= 
    \left\{ \begin{matrix} x & i=1\\ \tilde{d}^{\scalebox{0.4}{C}}_{i-1,M} & \mathrm{otherwise} \end{matrix} \right.\\
    \tilde{d}^{\scalebox{0.6}{C}}_{i,j} &= <g_{\phi_{i,j}}^{\scalebox{0.6}{C}}(\tilde{d}^{\scalebox{0.4}{C}}_{i,j-1}); z_i^{\scalebox{0.6}{BU}}> \text{ \textbf{for} } j=1,...,M\\
    y &= g_{\phi_{i,M+1}}^{\scalebox{0.4}{C}}(\tilde{d}^{\scalebox{0.4}{C}}_{i,M})\ ,
\end{align}
where $g_{\phi_{i,M+1}}^{\scalebox{0.4}{C}}$ is a final densely connected neural network layer, of the same dimension as the number of categories, and a Softmax activation function. The inference model is thereby factorized by:
\begin{align}
    q_{\phi}(\mathbf{z},y|x) = q_{\phi}(z_L|x,y,z_{<L}^{\scalebox{.6}{BU}}) q_{\phi}(y|x,z_{<L}^{\scalebox{.6}{BU}}) \prod_{i=1}^{L-1} q_{\phi}(z_i^{\scalebox{.6}{BU}}|x,z_{<i}^{\scalebox{.6}{BU}}) q_{\phi,\theta}(z_i^{\scalebox{.6}{TD}}|x,y,z_{<i}^{\scalebox{.6}{BU}},z_{>i}^{\scalebox{.6}{BU}}, z_{>i}^{\scalebox{.6}{TD}})\ .
\end{align}

\paragraph{Generative model.} For each stochastic latent variable, $\mathbf{z}$, and the observed variable $x$ in the generative model, as well as the TD path of the inference model, we add a conditional dependency on a categorical variable $y$:
\begin{align}
p_{\theta}(x, y, \mathbf{z}) = p_{\theta}(x|\mathbf{z},y) p_{\theta}(z_L) p_{\theta}(y) \prod_{i=1}^{L-1}p_{\theta}(z_i|z_{>i},y)\ .
\end{align}

\paragraph{Evidence lower bound.} In a semi-supervised learning problem, we have labeled data and unlabeled data which results in two formulations of the ELBO. The ELBO for labeled data points is given by:
\begin{align}
    \log p_{\theta}(x,y) \ge \mathbb{E}_{q_{\phi}(\mathbf{z}|x,y))}\left[\log \frac{p_{\theta}(x, y, \mathbf{z})}{q_{\phi,\theta}(\mathbf{z}|x,y)} \right] \equiv -\mathcal{F}(\theta, \phi)\ .
\end{align}
Since the classification model is not included in the above definition of the ELBO we add a classification loss term (a categorical cross-entropy), equivalent to the approach in \citep{Kingma14}:
\begin{align}
    \bar{\mathcal{F}}(\theta, \phi) = \bar{\mathcal{F}}(\theta, \phi) - \alpha \cdot \mathbb{E}_{q(z<L|x)} [\log q_{\phi}(y|x,z_{<L}^{\scalebox{.6}{BU}})] \ ,
\end{align}
where $\alpha$ is a hyperparameter that we define as in \citep{Maaloe2016}. For the unlabeled data points, we marginalize over the labels:
\begin{align}
    \log p_{\theta}(x) \ge \mathbb{E}_{q_{\phi}(\mathbf{z},y|x)}\left[\log \frac{p_{\theta}(x, y, \mathbf{z})}{q_{\phi,\theta}(\mathbf{z},y|x)} \right] \equiv -\mathcal{U}(\theta, \phi)\ .
\end{align}
The combined objective function over the labeled, $(x_l,y_l)$, and unlabeled data points, $(x_u)$, are thereby given by:
\begin{align}
    \mathcal{J}(\theta,\phi) = \sum_{x_l,y_l} \bar{\mathcal{F}}(\theta,\phi;x_l,y_l) + \sum_{x_u} \mathcal{U}(\theta,\phi;x_u) \ .
\end{align}

\newpage
\section{Additional Results}\label{app:additional_results}

\begin{table}[h!]
\begin{center}
\begin{small}
\begin{sc}
\caption{Test log-likelihood on dynamically binarized MNIST for different number of importance weighted samples. The finetuned models are trained for an additional number of epochs with no \textit{free bits}, $\lambda=0$.}\label{table:dyn_mnist}
\begin{tabular}{l c}
 & $-\log p(x)$ \\
 \hline
 \textit{Results with autoregressive components} \\
 DRAW+VGP \cite{Tran2016} & $<79.88$ \\
 IAFVAE \cite{Kingma2016} & $\leq 79.10$\\
 VLAE \cite{Chen2017} & $\leq 78.53$\\
 \hline
 \textit{Results without autoregressive components} \\
 IWAE \cite{Burda15} & $\leq 82.90$\\
 ConvVAE+HVI \cite{Salimans15} & $\leq 81.94$ \\
 LVAE \cite{Sonderby2016} & $\leq 81.74$ \\
 Discrete VAE \cite{Rolfe2017} & $\leq 80.04$ \\
 \abovespace
\textbf{\nm}, $\mathcal{L}_1$ & $\leq 80.60$ \\
\textbf{\nm}, $\mathcal{L}_1e3$ & $\leq 78.49$ \\
\textbf{\nm} finetuned, $\mathcal{L}_1$ & $\leq 80.06$ \\
\textbf{\nm} finetuned, $\mathcal{L}_{1e3}$ & $\leq 78.41$ \\
\hline
\end{tabular}%
\end{sc}
\end{small}
\end{center}
\end{table}

\begin{table}[h!]
\begin{center}
\begin{small}
\begin{sc}
\caption{Test log-likelihood on dynamically binarized OMNIGLOT for different number of importance weighted samples. The finetuned models are trained for an additional number of epochs with no \textit{free bits}, $\lambda=0$.}\label{table:omniglot}
\begin{tabular}{l c}
 & $-\log p(x)$ \\
\hline
\textit{Results with autoregressive components} \\
DRAW \citep{Gregor15} & $< 96.50$\\
ConvDRAW \cite{Gregor16} & $< 91.00$\\
VLAE \cite{Chen2017} & $\leq 89.83$\\
 \hline
 \textit{Results without autoregressive components} \\
 IWAE \cite{Burda15} & $\leq 103.38$\\
 LVAE \cite{Sonderby2016} & $\leq 102.11$\\
 DVAE \cite{Rolfe2017} & $\leq 97.43$\\
 \abovespace
\textbf{\nm}, $\mathcal{L}_1$ & $\leq 95.90$ \\
\textbf{\nm} finetuned, $\mathcal{L}_1$ & $\leq 93.54$ \\
\textbf{\nm} finetuned, $\mathcal{L}_{1e3}$ & $\leq 91.34$ \\
\hline
\end{tabular}
\end{sc}
\end{small}
\end{center}
\end{table}

\begin{table}[h!]
{\caption{Test log-likelihood on statically binarized Fashion MNIST for different number of importance weighted samples. The finetuned models are trained for an additional number of epochs with no \textit{free bits}, $\lambda=0$.}\label{table:fashion}}
\begin{center}
\begin{small}
\begin{sc}
\begin{tabular}{l c}
 & $-\log p(x)$ \\
\hline
\textbf{\nm}, $\mathcal{L}_1$ & $\leq 94.05$ \\
\textbf{\nm} finetuned, $\mathcal{L}_1$ & $\leq 93.54$ \\
\textbf{\nm} finetuned, $\mathcal{L}_{1e3}$ & $\leq 87.98$ \\
\hline
\end{tabular}%
\end{sc}
\end{small}
\end{center}
\end{table}

\begin{table}[h!]
\renewcommand\figurename{Table}
\begin{center}
\begin{small}
\begin{sc}
\caption{Test log-likelihood on ImageNet 32x32
for different number of importance weighted samples.
}\label{table:imagenet}
\begin{tabular}{l c}
 & bits/dim \\
\hline
\textit{With autoregressive components} \\
ConvDRAW {\scriptsize\cite{Gregor16}} & $<4.10$\\
PixelRNN {\scriptsize\cite{Oord2015}} & $= 3.63$\\
GatedPixelCNN {\scriptsize\cite{Oord16}} & $= 3.57$\\
 \hline
 \textit{Without autoregressive components} \\
RealNVP {\scriptsize\cite{Dinh16}} & $=4.28$\\
GLOW {\scriptsize\cite{Kingma18}} & $=4.09$\\
Flow++ {\scriptsize\cite{Ho19}} & $=3.86$\\
 \abovespace
\textbf{\nm}, $\mathcal{L}_1$ & $\leq 3.98$ \\
\textbf{\nm}, $\mathcal{L}_{1e3}$ & $\leq 3.96$ \\
\hline
\end{tabular}%
\end{sc}
\end{small}
\end{center}
	\vspace*{-0.6cm}
\end{table}

\begin{figure}[!h]
\begin{subfigure}{0.49\textwidth}
\centering
\includegraphics[width=1.\textwidth]{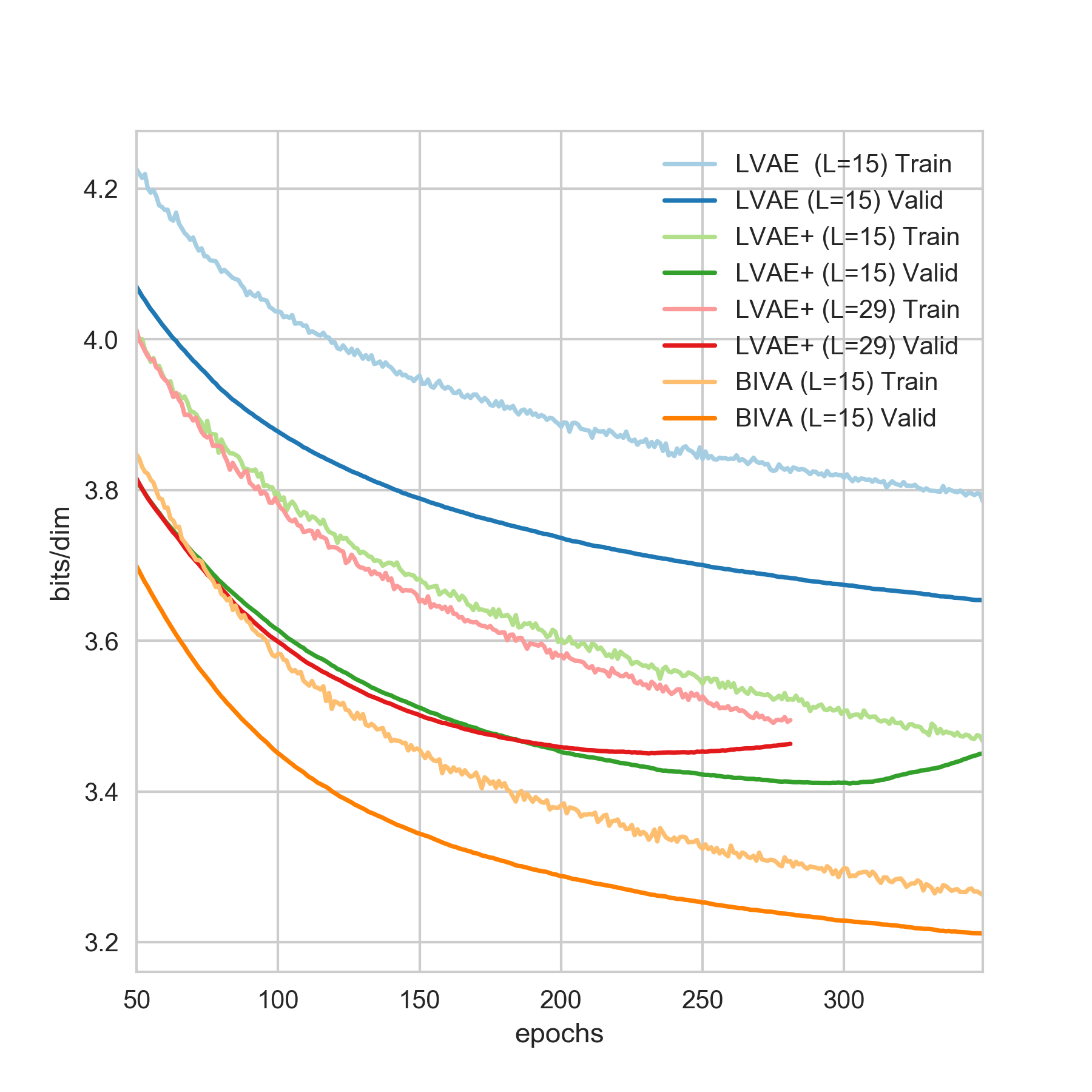}
\caption{$\mathcal{L}_1$ (bits/dim).}
\label{fig:convergence}
\end{subfigure}
\begin{subfigure}{0.49\textwidth}
\centering
\includegraphics[width=1.\textwidth]{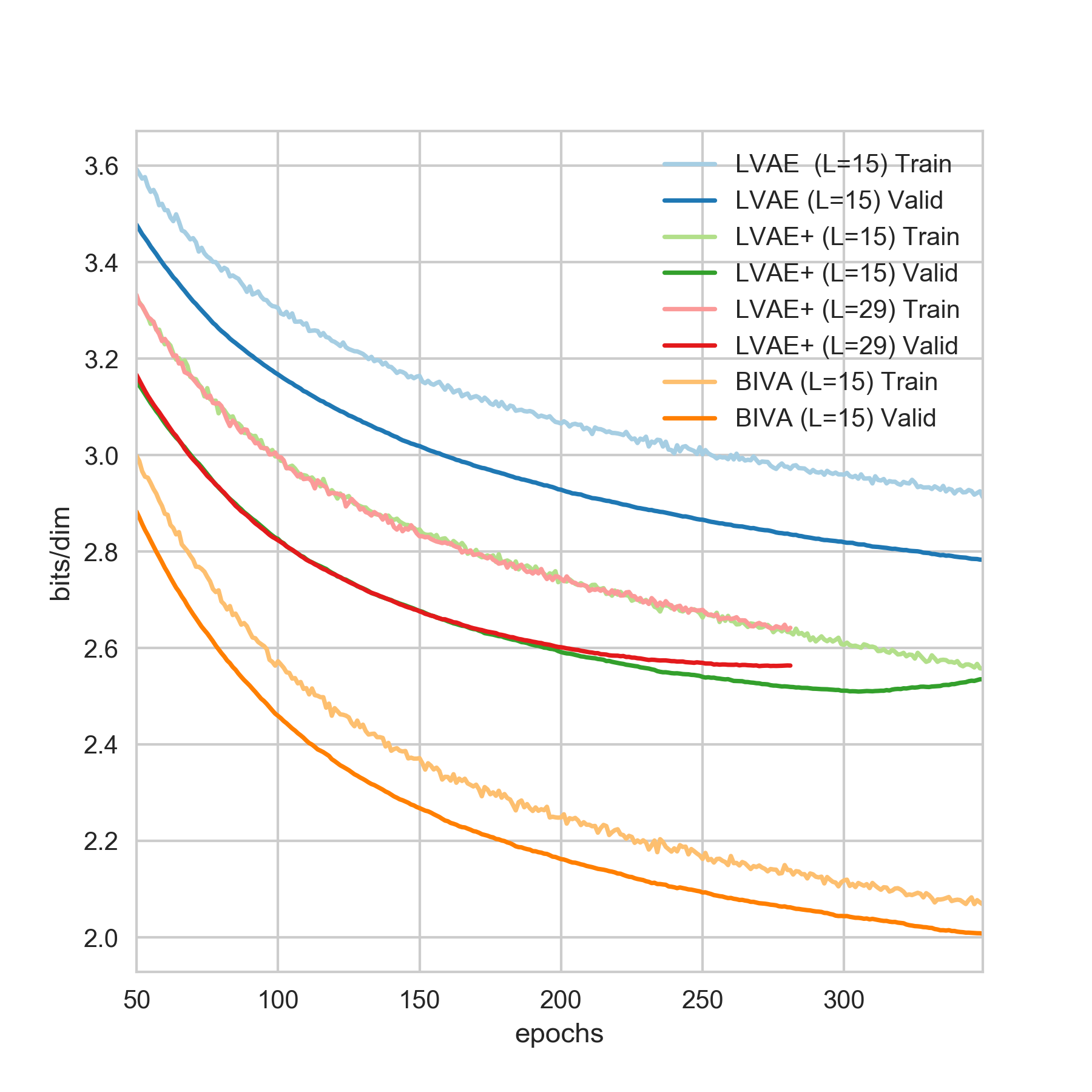}
\caption{$\log p_{\theta}(x|\mathbf{z})$ (bits/dim).}
\label{fig:convergence2}
\end{subfigure}
\caption{Convergence plot on CIFAR-10 training for the LVAE with $L=15$, the LVAE+ with $L=15$, the LVAE+ with $L=29$, and \nm with $L=15$. (a) shows the convergence of the 1 importance weighted ELBO, $\mathcal{L}_1$, calculated in bits/dim. (b) shows the convergence of the \textit{reconstruction loss}. The discrepancy between (a) and (b) is explained by the added cost from the stochastic latent variables, the Kullback-Leibler divergence $KL[p(\mathbf{z})||q(\mathbf{z}|x)]$.}
\end{figure}

\begin{figure}[!h]
  \begin{center}
    \includegraphics[width=0.48\textwidth]{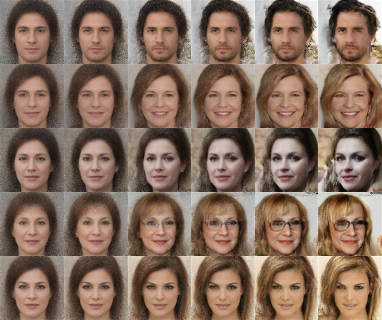}
  \end{center}
\caption{64x64 CelebA samples generated from a \nm with increasing levels of stochasticity in the model (going from close to the mode to the full distribution). In each column the latent variances are scaled with factors $0.1, 0.3, 0.5, 0.7, 0.9, 1.0$.
Images in a row look similar because they use the same Gaussian 
random noise $\mathbf{\epsilon}$ to generate the latent variables. \nm has $L=20$ stochastic latent layers connected by three layer ResNet blocks.}
\label{fig:celeba_var_gen}
\end{figure}

\begin{figure*}[!h]
\centering
\begin{subfigure}{0.49\textwidth}
\centering
\includegraphics[width=1.\textwidth]{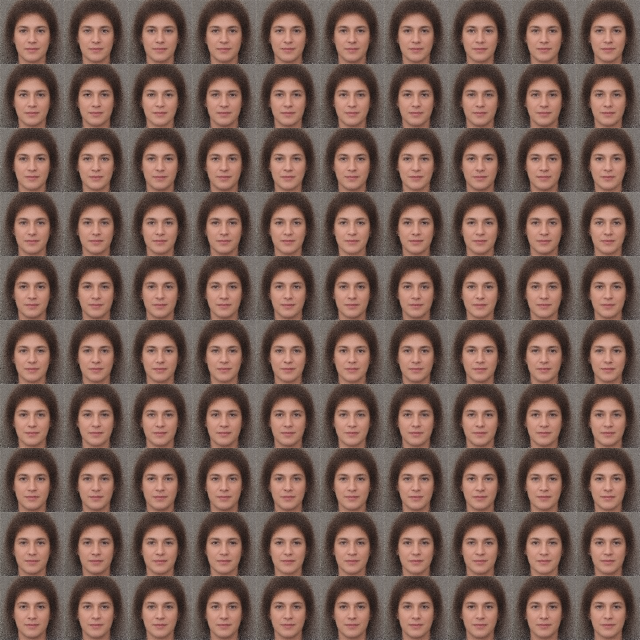}
\caption{$\sigma^2=0.01$}
\label{fig:temp001}
\end{subfigure}
\begin{subfigure}{0.49\textwidth}
\centering
\includegraphics[width=1.\textwidth]{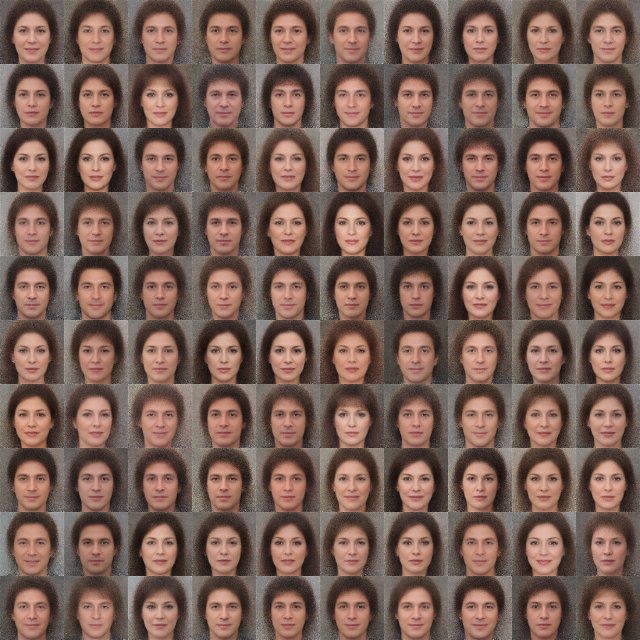}
\caption{$\sigma^2=0.1$}
\label{fig:temp01}
\end{subfigure}

\begin{subfigure}{0.49\textwidth}
\centering
\includegraphics[width=1.\textwidth]{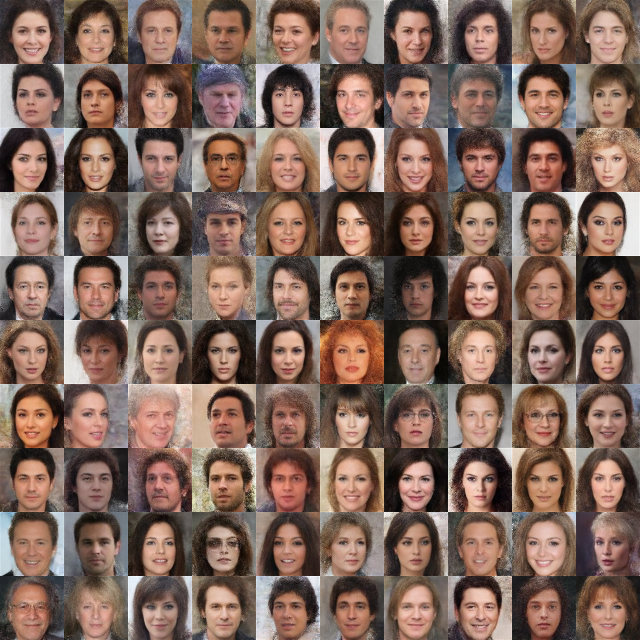}
\caption{$\sigma^2=0.5$}
\label{fig:temp01b}
\end{subfigure}
\begin{subfigure}{0.49\textwidth}
\centering
\includegraphics[width=1.\textwidth]{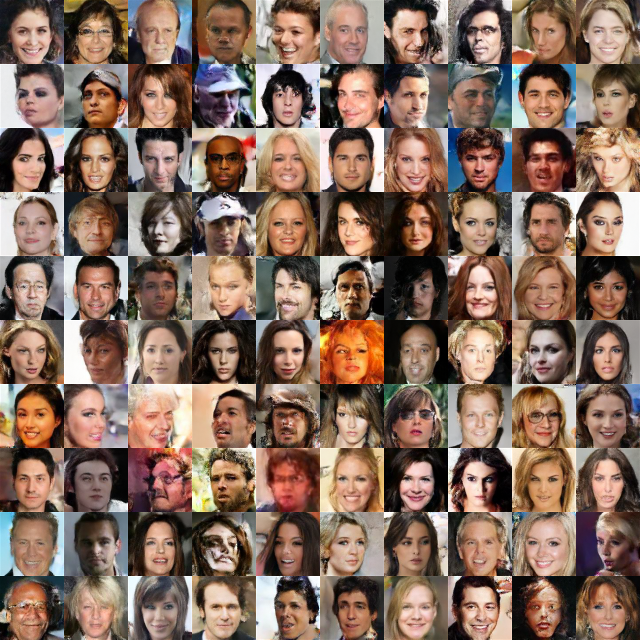}
\caption{$\sigma^2=1.0$}
\label{fig:temp001b}
\end{subfigure}
\caption{\nm $\mathcal{N}(0,\sigma^2)$ generations with varying $\sigma^2=0.01, 0.1, 0.5, 1.0$ for (a), (b), (c) and (d) respectively. We follow the same generating procedure of Figure \ref{fig:celeba_var_gen}. \nm has $L=20$ stochastic latent variables and is trained on the CelebA dataset, preprocessed to 64x64 images following \citep{Larsen16}. \nm achieves a $\mathcal{L}_1=2.48$ bits/dim on the test set. Close to the mode of the latent distribution there is very little variance in generated natural images. When we \textit{loosen} the samples towards the full distribution, $\sigma^2=1$, we can see how the generated images are adopting different styles and contexts.}
\end{figure*}

\begin{figure}[!h]
\centering
\includegraphics[width=.7\textwidth]{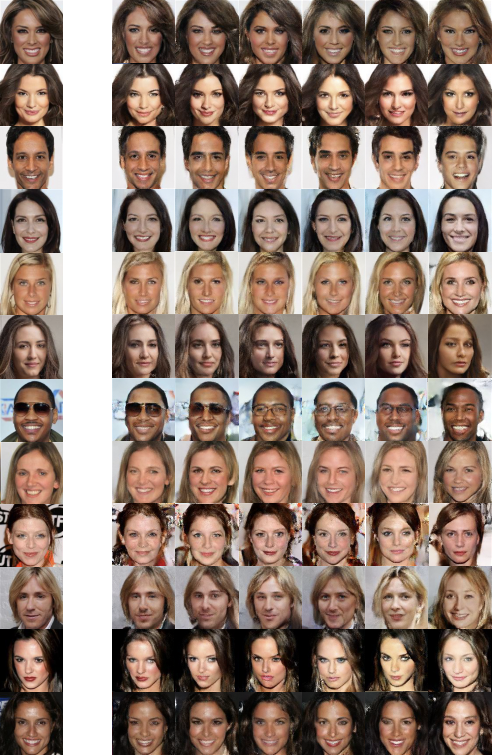}
\caption{\nm $L=20$ generations (right) from fixed $z_{>i}$ given an input image (left), for different layers throughout the stochastic variable hierarchy (from left to right $i=12,14,16,17,18,19$). The model is trained on CelebA, preprocessed to 64x64 images following \citep{Larsen16}. $z_{>i}$ are fixed by passing the original image through the encoder, after which $z_{\leq i}$ are sampled from the prior. When generating from a higher $z_i$ (columns) it is shown how the model has more \textit{freedom} to augment the input images. \nm achieves a $\mathcal{L}_1=2.48$ bits/dim on the test set.}
\end{figure}

\begin{figure}[!h]
\centering
\includegraphics[width=1.\textwidth]{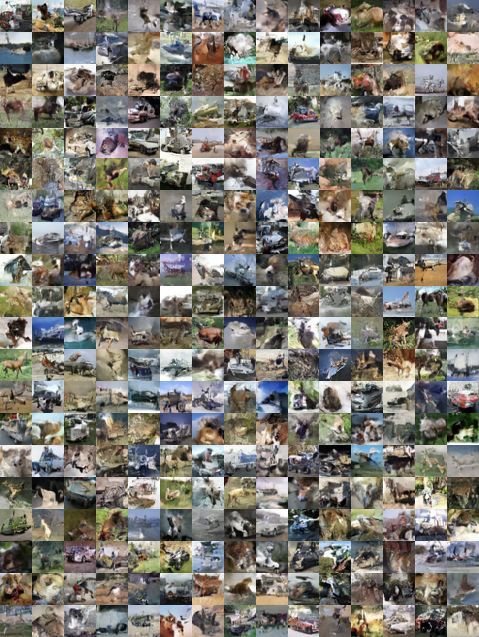}
\caption{\nm $\mathcal{N}(0,I)$ generations on a model trained on CIFAR-10. \nm has $L=15$ stochastic latent variables and achieves a $3.08$ bits/dim on the test set. The images are still not as sharp and coherent as the PicelCNN++ \citep{Salimans17} ($3.08$ vs. $2.92$), however, it does achieve to find coherent structure resembling the categories of the CIFAR-10 dataset.}
\end{figure}

\end{document}